\documentclass[conference]{IEEEtran}

\IEEEoverridecommandlockouts
\usepackage{cite}
\usepackage{amsmath,amssymb,amsfonts}
\usepackage{algorithmic}
\usepackage{graphicx}
\usepackage{textcomp}
\usepackage{xcolor}
\def\BibTeX{{\rm B\kern-.05em{\sc i\kern-.025em b}\kern-.08em
    T\kern-.1667em\lower.7ex\hbox{E}\kern-.125emX}}
    
\usepackage{multirow}
\usepackage{slashbox}
\usepackage{hyperref}

\newcommand{\ind}{\perp\!\!\!\!\perp} 

\begin{document}

\title{Equal Confusion Fairness: \\ Measuring Group-Based Disparities \\ in Automated Decision Systems}

% \author{\IEEEauthorblockN{Anonymous Author(s)}
% \IEEEauthorblockA{\textit{Anonymous Department} \\
% \textit{Anonymous Organization}\\
% Anonymous City, Country \\
% Anonymous email address}
% \and
% \and
% \and
% \and
% \and

\author{
    \IEEEauthorblockN{Furkan Gursoy, Ioannis A. Kakadiaris}
    \IEEEauthorblockA{\textit{Computational Biomedicine Lab} \\
\textit{Dept. of Computer Science} \\
\textit{University of Houston}\\
Houston, TX, USA
    \\ \{fgursoy, ioannisk\}@uh.edu}

\thanks{F. Gursoy and I. A. Kakadiaris, "Equal Confusion Fairness: Measuring Group-Based Disparities in Automated Decision Systems," 2022 IEEE International Conference on Data Mining Workshops (ICDMW), Orlando, FL, USA, 2022, pp. 137-146. \url{https://doi.org/10.1109/ICDMW58026.2022.00027}}
}

\maketitle

\begin{abstract}
As artificial intelligence plays an increasingly substantial role in decisions affecting humans and society, the accountability of automated decision systems has been receiving increasing attention from researchers and practitioners. Fairness, which is concerned with eliminating unjust treatment and discrimination against individuals or sensitive groups, is a critical aspect of accountability. Yet, for evaluating fairness, there is a plethora of fairness metrics in the literature that employ different perspectives and assumptions that are often incompatible. This work focuses on group fairness. Most group fairness metrics desire a parity between selected statistics computed from confusion matrices belonging to different sensitive groups. Generalizing this intuition, this paper proposes a new equal confusion fairness test to check an automated decision system for fairness and a new confusion parity error to quantify the extent of any unfairness. To further analyze the source of potential unfairness, an appropriate post hoc analysis methodology is also presented. The usefulness of the test, metric, and post hoc analysis is demonstrated via a case study on the controversial case of COMPAS, an automated decision system employed in the US to assist judges with assessing recidivism risks. Overall, the methods and metrics provided here may assess automated decision systems' fairness as part of a more extensive accountability assessment, such as those based on the system accountability benchmark. 
\end{abstract}

\begin{IEEEkeywords}
fairness, artificial intelligence, automated decision systems, algorithmic accountability, algorithm audit
\end{IEEEkeywords}

\section{Introduction}

Corresponding with the advances in artificial intelligence (AI) technology and the wider adoption of AI technologies by practitioners, automated decision systems (ADS) have begun to play an increasingly substantial role in assisting or making important decisions affecting human lives. Such decisions assisted by ADS include criminal recidivism risk assessment \cite{angwinMachineBias}, welfare fraud risk scoring \cite{vanbekkumDigitalWelfareFraud2021}, biometric recognition in law enforcement \cite{castelvecchiFacialRecognitionToo2020}, employment decisions \cite{raghavanMitigatingBiasAlgorithmic2020}, and visa application decisions \cite{mcleanDigitalJusticeAustralian2019}. Such uses of ADS are not free from issues such as bias and discrimination, and the referenced works include discussions on why those AI-based systems may be problematic.

The problematic applications of AI do not necessarily imply that all uses of ADS should be avoided. On the contrary, if their accountability is ensured, such systems may improve efficiency and effectiveness in many decision-making tasks. For instance, a systematic review of more than 50 papers found that majority of AI-enabled decision support systems improve patient safety outcomes in healthcare settings \cite{choudhuryRoleArtificialIntelligence2020}. However, the same study notes the lack of standardized benchmarks and homogeneous AI reporting. To this end, frameworks such as the system accountability benchmark \cite{gursoySystemCardsAIBased2022} aim to improve the standardization of AI accountability assessment and reporting within an exhaustive scheme. There are also legal and regulatory efforts to ensure accountability of ADS, mainly in the US \cite{clarkeAlgorithmicAccountabilityAct2022}, the EU \cite{EURLex52021PC0206EURLex}, and the UK \cite{GuidanceAIAuditing}.

Fairness is concerned with unjust outcomes for individuals or groups. Individual fairness postulates that similar persons should receive similar outcomes \cite{mehrabiSurveyBiasFairness2021}. Group fairness, on the other hand, is concerned with eliminating unjust outcomes based on sensitive group membership \cite{mehrabiSurveyBiasFairness2021}. Group fairness has been receiving increasing attention from researchers, practitioners, and legislators as many AI systems may exhibit bias based on race \cite{samoraniOverbookedOverlookedMachine2021}, gender \cite{pratesAssessingGenderBias2020}, age \cite{chuDigitalAgeismChallenges2022}, disability status \cite{whittakerDisabilityBiasAI}, political orientation \cite{petersAlgorithmicPoliticalBias2022}, and religion \cite{abidPersistentAntiMuslimBias2021}. This paper concentrates on group fairness.

There are multiple approaches and numerous notions and metrics for group fairness. These do not agree on a single fairness definition. This is so because fairness does not have a value-free definition, and different fairness approaches may adhere to different value principles. Consequently, the plethora of fairness metrics in the literature makes it challenging for practitioners to choose among many incompatible alternatives. It may also enable a "cherry-picking" behavior. This work aims to unify major fairness approaches and notions in a general but unique fairness assessment methodology and operationalize it to facilitate practical and effective use in the real world. The proposed methodology may also be employed to evaluate the group fairness elements included in larger accountability frameworks.

The main contributions of this paper can be enumerated as follows.
\begin{enumerate}
    \item Equal confusion fairness, a new group fairness notion, is introduced.
    \item The proposed notion is operationalized by designing appropriate testing and measurement processes.
        \begin{enumerate}
            \item An equal confusion test is designed to identify whether an ADS exhibits unfair behavior.
            \item A confusion parity error is proposed to quantify the extent of unfairness exhibited by the system.
            \item An appropriate methodology for the post hoc analysis is presented to identify the impacted groups and characterize the specific unfair behavior.
        \end{enumerate}
    \item A software program to assist with the analysis of equal confusion fairness is provided as an open-source tool.\footnote{The code and reproducibility files are made available at \url{https://github.com/furkangursoy/equalconfusion}.}
\end{enumerate}

The rest of the work is structured as follows. Section \ref{sec:rl} provides a comparative overview of the related work on group fairness. Section \ref{sec:ecf} presents the methods for the equal confusion test, confusion parity error, and the post hoc analysis. Section \ref{sec:case} demonstrates the applicability and usefulness of the proposed methods using a real-world dataset from an actual recidivism risk assessment tool that is employed in the US criminal justice system to assist judges in their decision-making. Final remarks and directions for future research are provided in Section \ref{sec:con}.

\section{Related Work}\label{sec:rl}

Albeit a relatively new topic, fairness in machine learning has seen a dramatic increase in publication numbers in recent years. Generally speaking, a distinction can be made between individual fairness and group fairness. While individual fairness focuses on whether similar individuals receive similar outcomes, group fairness focuses on whether the decisions are just for members of different groups on average. Usually, individual fairness notions employ distance functions to compute the similarity between individuals and the similarity between their respective outcomes. On the other hand, group fairness notions usually seek parity of selected statistics between different groups. Causality-based methods may be viewed as another stream. However, specific causality-based studies either focus on group fairness or individual fairness. This section summarizes major approaches to group fairness to provide a background for the methodology provided in the next section.

Three major approaches to group fairness exist: independence, separation, and sufficiency. All three are defined based on joint distributions of sensitive characteristics $s$, predictions $\hat{y}$, and ground truth values $y$. Independence requires that sensitive characteristics (e.g., race- or sex-based group memberships) and predictions are statistically independent. Separation requires that sensitive characteristics and predictions are conditionally independent given ground truth values. Sufficiency requires that sensitive characteristics and ground truth values are conditionally independent given predictions. The three approaches can be mathematically represented respectively as $s \ind \hat{y}$, $s \ind \hat{y} | y$, and $s \ind  y | \hat{y}$.

There is an abundance of fairness metrics in the literature. Mehrabi \textit{et al.} \cite{mehrabiSurveyBiasFairness2021} provided 10 widely used fairness measures. Makhlouf \textit{et al.} \cite{makhloufApplicabilityMachineLearning2021} presented 19 fairness measures, 16 of which are for group fairness. Castelnovo \textit{et al.} \cite{castelnovoClarificationNuancesFairness2022} and Verna and Rubin \cite{vermaFairnessDefinitionsExplained2018} presented 19 and 20 fairness measures, respectively. Fairness 360 toolkit by IBM \cite{bellamyAIFairness3602019} contains more than 70 fairness metrics as of 2022. 
Enumeration and the detailed investigation of those fairness metrics are beyond the scope of this work. Interested readers are encouraged to refer to the cited works and other surveys on the topic \cite{segalFairnessEyesData2021, chouldechovaSnapshotFrontiersFairness2020}. However, the following should be noted. Except for causality-based metrics, most group fairness metrics can be calculated from the confusion matrices belonging to different sensitive groups and many follow one of the three major approaches \cite{makhloufApplicabilityMachineLearning2021, razGroupFairnessIndependence2021}.

Confusion matrices tabulate the relationship between $\hat{y}$ and $y$, providing information on the type of errors made by a classifier. For binary classification, a confusion matrix consists of four cells, as shown in Table \ref{tab:confMatrix}. The cells contain the frequencies for true positives ($TP$), false positives ($FP$), false negatives ($FN$), and true negatives ($TN$). From a confusion matrix, additional statistics can be defined. Precision is defined as the fraction of actual positives among all positive predictions. Negative predictive value is defined as the fraction of actual negatives among all negative predictions. Recall is defined as the fraction of predicted positives among all actual positives. Specificity is defined as the fraction of predicted negatives among all actual negatives. Their mathematical definitions are given below. Any three of the four are necessary and sufficient to compute the fourth and to fully identify the distribution of the confusion matrix:
\begin{itemize}
    \item Precision: $TP / (TP + FP)$,
    \item Negative Predictive Value: $TN / (TN + FN)$,
    \item Recall:  $TP / (TP + FN)$, and
    \item Specificity: $TN / (TN + FP)$.
\end{itemize}

%%%%%%%%%%%%%%%%%%%%%%%%%%%%%%%%%%%%%%%%%%%%%%%%%%%%
\begin{table}[t]
\caption{Confusion matrix definition.}
\centering
\begin{tabular}{cc|cc|}
\cline{3-4}
                                                 &     & \multicolumn{2}{c|}{Actual}      \\ \cline{3-4} 
                                                 &     & \multicolumn{1}{c|}{$+$}  & $-$  \\ \hline
\multicolumn{1}{|c|}{\multirow{2}{*}{Predicted}} & $+$ & \multicolumn{1}{c|}{$TP$} & $FP$ \\ \cline{2-4} 
\multicolumn{1}{|c|}{}                           & $-$ & \multicolumn{1}{c|}{$FN$} & $TN$ \\ \hline
\end{tabular}
\label{tab:confMatrix}
\end{table}
% Overall confusion
%%%%%%%%%%%%%%%%%%%%%%%%%%%%%%%%%%%%%%%%%%%%%%%%%%%%

In relation to confusion matrices, the three major fairness approaches require the following respective quantities to be on par across sensitive groups \cite{razGroupFairnessIndependence2021}:

\begin{itemize}
    \item Independence: $(TP + FP)/((TP + FP + FN + TN)$,
    \item Sufficiency:  $TP / (TP + FP)$ and $TN / (TN + FN)$ (i.e.,  precision and negative predictive value, respectively), and
    \item Separation: $TN / (TN + FP)$  and $TP / (TP + FN)$ (i.e., specificity and recall, respectively).
\end{itemize}

When sufficiency and separation are known, the distribution of the confusion matrix becomes known. Hence, independence may also be computed. Moreover, once the distribution is known, other fairness metrics based on confusion matrices may also be computed. The case of all three fairness approaches being satisfied is known as total fairness \cite{berkFairnessCriminalJustice2021}. However, it is not possible to satisfy all three at the same time except in specific cases \cite{barocasFairnessMachineLearning2019, razGroupFairnessIndependence2021}.

Although their simultaneous satisfaction is rarely observed outside rhetorical cases \cite{berkFairnessCriminalJustice2021}, all three fairness approaches are sought to be satisfied as much as possible. This paper argues that while the impracticality regarding the simultaneous and perfect satisfaction of the three approaches should be acknowledged, practitioners should strive to achieve the best possible overall performance in all three. This would also prevent "cherry-picking" among more confined fairness metrics when evaluating an ADS for fairness. Therefore, this paper argues that the distribution of confusion matrices, from which most group fairness metrics are computed, should be on par across different groups. As the specific source(s) of a potential unfairness result would not be immediately apparent, any unfairness result should be followed up by an appropriate post hoc analysis that seeks to reveal and characterize the inequalities between the confusion matrices.

Another noteworthy and relevant concept is intersectional fairness \cite{fouldsIntersectionalDefinitionFairness2020}. Intersectionality is a framework to study how overlapping identities may create different inequities in the sense that the sum is more than the parts. Thus,  intersectional fairness requires the analysis of intersectional groups (e.g., Hispanic females) rather than isolated analyses of, for instance, race and sex. The intersectional approach also limits fairness gerrymandering \cite{kearnsPreventingFairnessGerrymandering2018} where a system appears fair at a group level but is not fair at a subgroup level.

\section{Equal Confusion Fairness}\label{sec:ecf}

\subsection{Notation}

Scalar values are denoted by lower case letters (e.g., $a$).
Vectors are denoted by boldface lowercase letters (e.g., $\mathbf{a}$). The $i^{th}$ element of $\mathbf{a}$ is denoted by $\mathbf{a}_i$.
Matrices are denoted by boldface uppercase letters (e.g., $\mathbf{A}$). 
The $i^{th}$ row vector and $j^{th}$ column vector of $\mathbf{A}$ are denoted by $\mathbf{A}_{i*}$ and $\mathbf{A}_{*j}$, respectively.
The entry at the intersection of $i^{th}$ row and $j^{th}$ column of $\mathbf{A}$ is denoted by $\mathbf{A}_{ij}$.  
The real value space, nonnegative integer space, and categorical value space are denoted respectively by $\mathbb{R}$, $\mathbb{Z}^+$, and $\mathbb{S}$.
A vector of categorical values with size $n$ is denoted as  $\mathbf{a} \in \mathbb{S}^{n}$. A non-negative integer-valued matrix with $n$ rows and $m$ columns is denoted as $\mathbf{A} \in \mathbb{Z^+}^{n \times m}$.

\subsection{Problem Definition}

This paper proposes an equal confusion approach to investigate the fairness of a decision system, given the following:

\begin{itemize}

    \item a matrix $\mathbf{X}$ that represents $n$ humans and $m$ features where $\mathbf{X} \in \mathbb{(R \cup S)}^{n \times m}$,

    \item a vector $\mathbf{s}$ that represents the sensitive group memberships for the $n$ humans where $\mathbf{s} \in \mathbb{S}^{n}$  regardless of whether $\mathbf{s} \ind \mathbf{X}$ in general,

    \item a decision system $f: \mathbf{X} \rightarrow \mathbf{\hat{y}}$,
    
    \item decision outputs $\mathbf{\hat{y}}$ where $\mathbf{\hat{y}} \in \mathbb{S}^{n}$, and
    
    \item corresponding ground truth values $\mathbf{y}$ where $\mathbf{y} \in \mathbb{S}^{n}$.
    
\end{itemize}

Equal confusion fairness requires the confusion matrices to have the same distribution across all sensitive groups. To this end, first, a statistical test is presented to determine whether a decision system is fair or not. Second, a fairness metric is presented to measure the extent of unfairness, if any. Third, a post hoc test is presented to detect the differences in specific sensitive groups and specific decision system behavior contributing to unfairness, if any.

\subsection{Equal Confusion Test}
To determine whether a decision system is fair or not, equal confusion fairness investigates the relation between sensitive groups and outcome groups. 
Usually, $\mathbf{s}$ represents protected groups such as those based on gender and race. 
The pair $\{\mathbf{\hat{y}}, \mathbf{y}\}$ represents outcome groups. Specifically, the unique value pairs in $\{\mathbf{\hat{y}}, \mathbf{y}\}$ correspond to the cells in the confusion matrix. For instance, in the case of a binary decision problem, outcome groups are true positive, false positive, true negative, and false negative. 

The equal confusion test employs Pearson's chi-squared test of independence to test the relationship between $\mathbf{s}$ and $\{\mathbf{\hat{y}}, \mathbf{y}\}$. The relevant null and alternate hypotheses for Pearson's chi-squared test of independence are as follows.

$\mathbf{H_0}$:  $\mathbf{s}$ and $\{\mathbf{\hat{y}}, \mathbf{y}\}$ are independent.

$\mathbf{H_A}$:  $\mathbf{s}$  and $\{\mathbf{\hat{y}}, \mathbf{y}\}$ are dependent.

\begin{figure}[t]
\centerline{\includegraphics[width=0.475\textwidth]{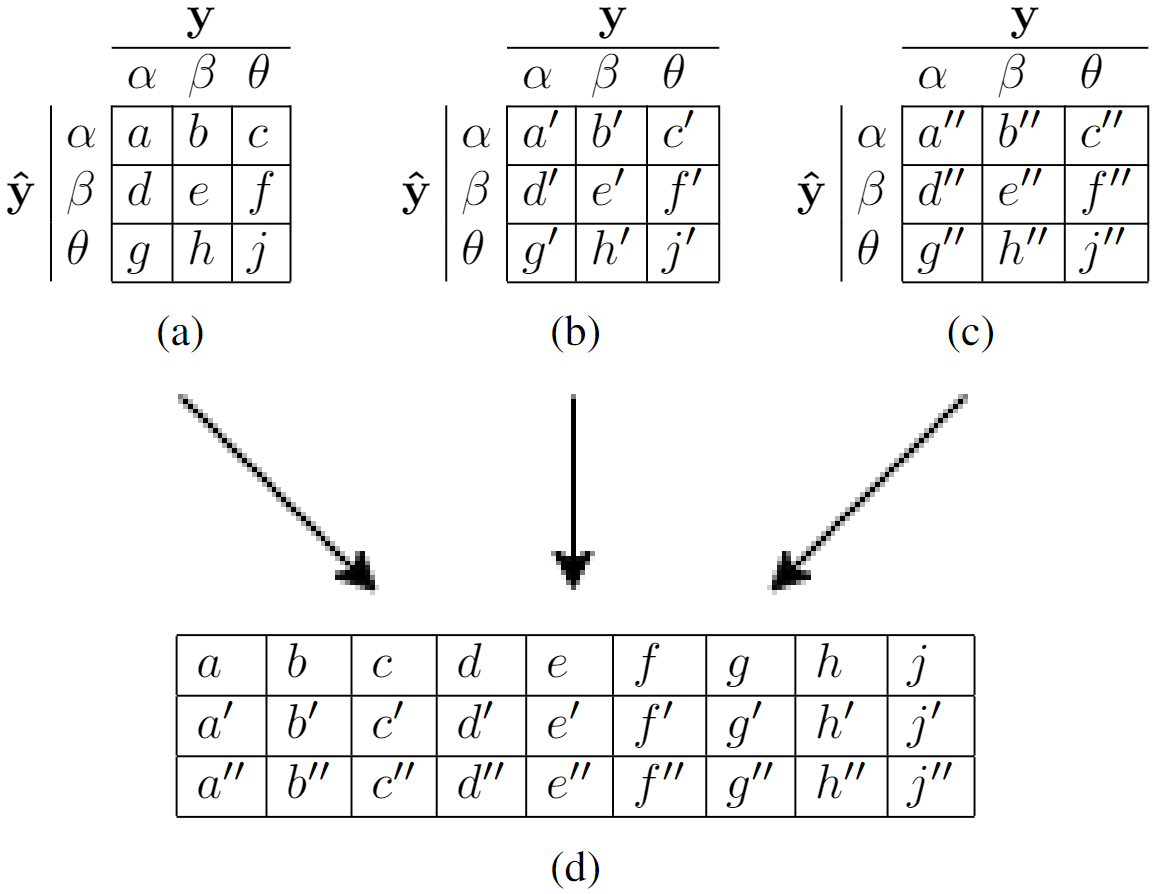}}
\caption{Confusion matrices to contingency matrix. (a) $\mathbf{C}^1$, (b) $\mathbf{C}^2$, (c) $\mathbf{C}^3$, (d) $\mathbf{O}$.}
\label{fig:contingency}
\end{figure}

The test requires a contingency matrix $\mathbf{O} \in \mathbb{Z^+}^{q \times r}$ where $q$ is the number of sensitive groups and $r$ is the number of outcome groups (i.e., the number of cells in the confusion matrix). 
The contingency matrix cross-tabulates the observed frequencies for sensitive groups and outcome groups.
Fig. \ref{fig:contingency} illustrates the generation of the contingency matrix from the set of confusion matrices for a decision system with three possible outputs (i.e., $\mathbf{\hat{y}}_{i}, \mathbf{y}_{i} \in \{\alpha, \beta, \theta\}$) and three sensitive groups. For this system, $q=3$ and $r=9$. 
Figs. \ref{fig:contingency}a, \ref{fig:contingency}b, and \ref{fig:contingency}c represent the confusion matrices $\mathbf{C}^1$, $\mathbf{C}^2$, and $\mathbf{C}^3$, respectively, corresponding to the three sensitive groups. 
Consequently, for instance, the cell value $b^\prime$ corresponds to the number of people (i) who belong to the second sensitive group, (ii) for whom the decision system produced the label $\mathbf{\hat{y}}_{i}=\alpha$, and (iii) whose th label is $\mathbf{y}_{i}=\beta$. 
Each confusion matrix is flattened to obtain a single vector. 
The obtained vectors are stored in the rows of the contingency matrix (Fig. \ref{fig:contingency}d). Hence, $\mathbf{O}_{i*}$ is equivalent to $\mathbf{C}^i$. 
Therefore, sensitive groups and outcome groups are represented respectively in the rows and columns of the contingency matrix.

After establishing $\mathbf{O}$, the expectation matrix $\mathbf{E}$ is computed. The matrix $\mathbf{E}$ has the same shape as $\mathbf{O}$ and represents the case of independence between $\mathbf{s}$ and $\{\mathbf{\hat{y}}, \mathbf{y}\}$, the expected frequencies under the null hypothesis. The values of its entries, $\mathbf{E}_{ij}$, are computed as shown in Eq. \ref{eq:exp}. 

\begin{equation}
  \mathbf{E}_{ij} = \frac{\sum\limits_{k=1}^{q}{\mathbf{O}_{kj}} \sum\limits_{l=1}^{r}{\mathbf{O}_{il}}} {\sum\limits_{k=1,l=1}^{q,r}{\mathbf{O}_{kl}}} 
\label{eq:exp}\end{equation}

Then, the chi-squared statistic $\chi^2$, the sum of normalized squared differences between the observed and expected values, is computed as shown in Eq. \ref{eq:chi}.

\begin{equation}
\chi^2 = \sum_{i=1}^q \sum_{j=1}^r \frac{(\mathbf{O}_{ij} - \mathbf{E}_{ij})^2}{\mathbf{E}_{ij}}
\label{eq:chi}\end{equation}

To evaluate the significance level for Pearson's chi-squared test of independence, the corresponding $p$ value can be obtained from the chi-squared distribution with $(q-1)(r-1)$ degrees of freedom. If it is found as statistically significant (e.g., $p<0.01$), the null hypothesis is rejected and the strength of the association between $\mathbf{s}$ and $\{\mathbf{\hat{y}}, \mathbf{y}\}$ is investigated next.

\subsection{Confusion Parity Error}
The confusion parity error is equivalent to Cramer's V \cite{cramerMathematicalMethodsStatistics1946} computed on $\mathbf{O}$. It is a measure of the association between two categorical variables based on the chi-squared statistic. It generalizes the Matthews correlation coefficient \cite{matthewsComparisonPredictedObserved1975} beyond binary variables, which is otherwise only applicable to two-by-two contingency matrices (i.e., extending it for $r>4$). Cramer's V, denoted by $\phi$, is computed as shown in Eq. \ref{eq:phi}. 

\begin{equation}
\phi = \sqrt{\frac{\chi^{2} / n}{\min (q-1, r-1)}}
\label{eq:phi}
\end{equation}

Its range is $[0,1]$ irrespective of the shape of $\mathbf{O}$. The value $0$ corresponds to no association and $1$ corresponds to complete association. The lower bounds of $\phi$ for determining small, moderate, or strong association strength are presented in Table \ref{tab:phinterpretation} following the recommendations provided by \cite{cohenStatisticalPowerAnalysis}. However, interpreting such effect sizes requires caution and may depend on the context \cite{fergusonEffectSizePrimer2009}.

\subsection{Post hoc Fairness Analysis}

Pearson's chi-squared test of independence is omnibus. That is, it is a global test that does not reveal the specific source of a statistically significant result \cite{Sharpe2015YourCT}. In the simplest case of a binary classification with only two sensitive groups (i.e., $q=2$ and $r=4$), a statistically significant fairness test result does not reveal which cells of the original confusion matrix (i.e., the cells that denote true positive, false positive, true negative, and false negative) contribute towards the statistically significant result. In the case of more than two sensitive groups (i.e., $q > 2$), a statistically significant test result does not reveal among which groups the identified discrepancy exists. Here, a suitable post hoc analysis method is presented to identify the contingency matrix cells contributing to the unfairness determined by the fairness test.

Adjusted standardized residual $\mathbf{R}_{ij}$  for a specific cell $\mathbf{O}_{ij}$ is computed by finding the difference between the observed and the expected value and then normalizing this value with an appropriate adjustment and standardization \cite{habermanAnalysisResidualsCrossClassified1973}. Equation \ref{eq:mek} presents adjusted standardized residual $\mathbf{R}_{ij}$ that corresponds to the cell $\mathbf{O}_{ij}$. 

\begin{equation}
  \mathbf{R}_{ij} = \frac{\mathbf{O}_{ij} -  \mathbf{E}_{ij} }{\sqrt{\mathbf{E}_{ij}  (1-\frac{\sum\limits_{l=1}^{r}{\mathbf{O}_{il}}}{\sum\limits_{k=1,l=1}^{q,r}{\mathbf{O}_{kl}}})
  (1-\frac{\sum\limits_{k=1}^{q}{\mathbf{O}_{kj}}}{\sum\limits_{k=1,l=1}^{q,r}{\mathbf{O}_{kl}}})
  }} 
\label{eq:mek}\end{equation}

The residual $\mathbf{R}_{ij}$ is then tested against the standard normal distribution at an appropriate significance level \cite{Agresti2007}. It is suggested in the literature to either apply a Bonferroni correction based on the number of cells in the contingency matrix \cite{macdonaldTypeErrorRate2000, Sharpe2015YourCT} or evaluate the statistical significance at a stricter level \cite{Sharpe2015YourCT}. For a desired statistical significance level of $95\%$, the appropriate \textit{p-value} would be $\frac{0.05}{qr}$ after the Bonferroni correction instead of $0.05$. A more stringent \textit{p-value} such as $0.001$ is recommended in the latter. Employing the \textit{p-value} of $0.001$, $\mathbf{R}_{ij}$ values less than $-3.29$ indicate a smaller value of $\mathbf{O}_{ij}$ than expected. $\mathbf{R}_{ij}$ values more than $3.29$ indicate that $\mathbf{O}_{ij}$ is higher than expected. As the expectation reflects no discrepancy between the sensitive groups in the confusion matrix, such deviations reveal the specific sources of unfairness.

\begin{table}[t]
 \caption{Interpreting Cramer's V ($\phi$). Values are computed based on the effect size index provided in \cite{cohenStatisticalPowerAnalysis}.}

    \centering
\begin{tabular}{rr|rrrrrrr|}
\cline{3-9}
\multicolumn{1}{l}{}                            & \multicolumn{1}{l|}{} & \multicolumn{7}{c|}{$min(q, r)$}                                                                                                                                                          \\ \cline{3-9} 
\multicolumn{1}{c}{}                            & \multicolumn{1}{c|}{} & \multicolumn{1}{c|}{4}   & \multicolumn{1}{c|}{5}   & \multicolumn{1}{c|}{6}   & \multicolumn{1}{c|}{7}   & \multicolumn{1}{c|}{8}   & \multicolumn{1}{c|}{9}   & \multicolumn{1}{c|}{10} \\ \hline
\multicolumn{1}{|r|}{\multirow{3}{*}{strength}} & small                 & \multicolumn{1}{r|}{.06} & \multicolumn{1}{r|}{.05} & \multicolumn{1}{r|}{.04} & \multicolumn{1}{r|}{.04} & \multicolumn{1}{r|}{.04} & \multicolumn{1}{r|}{.04} & .03                     \\ \cline{2-9} 
\multicolumn{1}{|r|}{}                          & moderate              & \multicolumn{1}{r|}{.17} & \multicolumn{1}{r|}{.15} & \multicolumn{1}{r|}{.13} & \multicolumn{1}{r|}{.12} & \multicolumn{1}{r|}{.11} & \multicolumn{1}{r|}{.11} & .10                     \\ \cline{2-9} 
\multicolumn{1}{|r|}{}                          & strong                & \multicolumn{1}{r|}{.29} & \multicolumn{1}{r|}{.25} & \multicolumn{1}{r|}{.22} & \multicolumn{1}{r|}{.20} & \multicolumn{1}{r|}{.19} & \multicolumn{1}{r|}{.18} & .17                     \\ \hline
\end{tabular}
    \label{tab:phinterpretation}
\end{table}

\subsection{Complexity Analysis}
Computing the contingency matrix (Fig. \ref{fig:contingency}) has the computational complexity of $O(n)$ where $n$ is the number of humans. Computational complexity of computing expected values (Eq. \ref{eq:exp}), computing $\chi^2$ (Eq. \ref{eq:chi}), and computing adjusted standardized residuals (Eq. \ref{eq:mek}) is $O(qr)$. The computational complexity for computing Cramer's V is $O(1)$. In most realistic cases,  $q$ and $r$ are very small. The computational complexity of $O(n)$ indicates a linear time. The space complexity is $O(qr)$. Hence, the presented methods are highly scalable.

The presented methodology has specific sample size requirements, as will be stated next. Therefore, sample complexity is a more constraining factor in comparison to time and space complexities.

\subsection{Scope, Discussion, and Limitations}

A list of considerations is provided below to clarify the scope of the applicability of the presented techniques and their limitations.

\begin{itemize}
    \item The approach is applicable for binary and multi-class classification tasks. It is not applicable for regression tasks.
    \item Reliable and unbiased ground truth labels are required.
    \item A representative and acceptable test set is required. In practice, the representativeness of a test may not be determined with absolute certainty. Therefore, an ongoing fairness assessment that utilizes the data collected via the system's real-world use is highly recommended as part of a larger monitoring strategy.
    \item The test set and the frequencies in each cell should be sufficiently large to allow a reliable statistical analysis. Cochran \cite{cochranMethodsStrengtheningCommon1954} recommends for Pearson's chi-squared test of independence that expected frequencies in the contingency matrix should be (i) at least five for at least 80\% of the cells and (ii) at least one in all cells.
    \item The presented fairness approach is rather strict and forces independence, sufficiency, and separation, which can be simultaneously maximized only in very restrictive cases.
    \item If there is more than a single type of sensitive group (e.g., when both gender and race need to be considered), the test can be repeated separately for gender groups and race groups. If it is desired to compare intersectional groups, such groups may be created from race and gender (e.g., black men, black women, white men, white women, and so on). Therefore, the presented techniques are also suitable for intersectional perspectives.
    \item If there is more than one dependent variable, the test can be repeated for each dependent variable separately. Alternatively and additionally, it can be repeated for intersections of the dependent variables, similar to the intersectional groups.
    \item In certain cases, the fairness test and post hoc analysis results may not agree. For instance, an unfairness detected by the fairness test may not be traced to individual cells by the post hoc analysis. Such a result may be due to the statistical significance of a combination of multiple cells where no single cell can be individually identified as statistically significant \cite{coxPostHocPairWise1993}.
\end{itemize}

\section{Case Study}\label{sec:case}

Correctional Offender Management Profiling for Alternative Sanctions (COMPAS) is a commercially developed automated decision-aiding tool for risk assessment in criminal justice. It has been used in several US states, including Florida, New York, Wisconsin, and California \cite{kirkpatrickItNotAlgorithm2017}. It can produce risk scores for recidivism, violent recidivism, and failure to appear \cite{mattuHowWeAnalyzed}. It uses a proprietary methodology, and the underlying computations are made available neither to the defendant nor to the court \cite{israniAlgorithmicDueProcess}.

In 2016, in a case brought by a defendant against the State, the Wisconsin Supreme Court decided that the use of COMPAS by a court did not violate the defendant's due process rights \cite{n.w.2d749StateLoomis}. Later in 2017, the Supreme Court of the United States denied an appeal by the defendant \cite{LoomisWisconsin}. Nevertheless, the use of COMPAS remained controversial, with several studies performed on the subject \cite{angwinMachineBias, dieterichCOMPASRiskScales2016, washingtonHowArgueAlgorithm2019, jacksonSettingRecordStraight2020}.

\subsection{Methodology}
The dataset used in this paper is published \cite{propublicaDataAnalysisMachine2022} alongside the original ProPublica story \cite{angwinMachineBias} that attracted widespread attention to the subject. The dataset is originally obtained via public information requests in Broward County, Florida. It contains 18,610 people who were scored in 2013 and 2014. COMPAS can be used in different stages in the criminal justice system, including parole and probation. However, this particular county primarily uses it at the pretrial stage \cite{mattuHowWeAnalyzed}, for which there are a total of 11,757 people.

At the pretrial stage, COMPAS produces scores including recidivism risk and violent recidivism risk. This case study focuses on violent recidivism which consists of murder and nonnegligent manslaughter, forcible rape, robbery, and aggravated assault by the FBI definition \cite{ViolentCrime}. COMPAS scores are integers from 1 to 10, where scores from 1 to 4 correspond to low risk, 5 to 7 correspond to medium risk, and scores above 7 correspond to high risk. According to the practitioner guide for COMPAS \cite{northpointePractitionersGuideCOMPAS2012}, medium and high scores receive more interest from supervision agencies. Therefore, in line with the original ProPublica analysis \cite{mattuHowWeAnalyzed}, medium and high risk are indicated as a positive prediction for recidivism. According to its practitioner guide \cite{northpointePractitionersGuideCOMPAS2012}, the COMPAS recidivism score predicts the risk of recidivism in the next two years. Via the data collected from public criminal records, the dataset also contains information on whether the defendant is charged with a violent criminal offense within two years after the original COMPAS screening \cite{angwinMachineBias}.

Following the same procedure as ProPublica \cite{mattuHowWeAnalyzed}, filtering is performed to remove (i) cases where the COMPAS assessment date is not within 30 days of arrest or charge dates, (ii) cases where COMPAS assessment is not found, and (iii) cases where the defendant did not have at least two years outside a correctional facility. The final dataset contains 4,020 COMPAS cases. For each case, the following information is available:
\begin{itemize}
    \item Sex: Female, Male;
    \item Race: African-American, Asian, Caucasian, Hispanic, Native American, Other;
    \item Predicted violent recidivism: 0--Non-risky, 1--Risky; and
    \item Actual violent recidivism: 0--Recidivist, 1--Non-recidivist.
\end{itemize}

Initially, descriptive statistics are explored to have a general understanding of the data. Then, for sensitive groups based on sex, race, and their intersections, fairness assessment studies are conducted. The equal confusion test is used to check whether the system exhibits unfair behavior, followed by the confusion parity error to measure the magnitude of unfairness. Finally, a post hoc fairness analysis is performed to reveal the specific characteristics of the unfairness and impacted groups. These analyses are supported by a set of tables presenting information on observed and expected values for the contingency matrices, which by construction contains information on their constituent confusion matrices.

\subsection{Findings}

\textit{Descriptive Statistics.} Table \ref{tab:dist} provides the distribution of the cases over race, gender, and their intersection. A large majority of the cases are male. However, females represent one-fifth of the cases, with nearly 900 cases. Caucasians and African-Americans together constitute 84\% of all cases while Asians and Native Americans collectively account for only less than 1\% with only 33 cases. The low number of cases for these two groups indicates that it will be very difficult, if not impossible, to achieve any statistically significant results for them.

\begin{table}[b]
\caption{Race and gender distribution of the cases.}
\centering
\begin{tabular}{r|r|r|r|r}
\cline{2-5}
                                       & \multicolumn{1}{c|}{Female} & \multicolumn{1}{c|}{Male} & \multicolumn{1}{c|}{Total} & \multicolumn{1}{c|}{\%}    \\ \hline
\multicolumn{1}{|r|}{African-American} & 393                         & 1,525                     & 1,918                      & \multicolumn{1}{r|}{48\%}  \\ \hline
\multicolumn{1}{|r|}{Asian}            & 1                           & 25                        & 26                         & \multicolumn{1}{r|}{1\%}   \\ \hline
\multicolumn{1}{|r|}{Caucasian}        & 336                         & 1,123                     & 1,459                      & \multicolumn{1}{r|}{36\%}  \\ \hline
\multicolumn{1}{|r|}{Hispanic}         & 61                          & 294                       & 355                        & \multicolumn{1}{r|}{9\%}   \\ \hline
\multicolumn{1}{|r|}{Native American}  & 0                           & 7                         & 7                          & \multicolumn{1}{r|}{0\%}   \\ \hline
\multicolumn{1}{|r|}{Other}            & 50                          & 205                       & 255                        & \multicolumn{1}{r|}{6\%}   \\ \hline
\multicolumn{1}{|r|}{Total}            & 841                         & 3,179                     & 4,020                      & \multicolumn{1}{r|}{100\%} \\ \hline
\multicolumn{1}{|r|}{\%}               & 21\%                        & 79\%                      & 100\%                      &                            \\ \cline{1-4}
\end{tabular}
\label{tab:dist}
\end{table}

%%%%%%%%%%%%%%%%%%%%%%%%%%%%%%%%%%%%%%%%%%%%%%%%%%%%
\begin{table}[b]
\caption{Overall confusion matrix for the cases.}
\centering
\begin{tabular}{rr|rrrr}
\cline{3-6}
                                                 &       & \multicolumn{4}{c|}{Actual}                                                                                      \\ \cline{3-6} 
                                                 &       & \multicolumn{1}{c|}{$+$}  & \multicolumn{1}{c|}{$-$}   & \multicolumn{1}{c|}{Total} & \multicolumn{1}{c|}{\%}    \\ \hline
\multicolumn{1}{|r|}{\multirow{4}{*}{Predicted}} & $+$   & \multicolumn{1}{r|}{346}  & \multicolumn{1}{r|}{761}   & \multicolumn{1}{r|}{1,107} & \multicolumn{1}{r|}{28\%}  \\ \cline{2-6} 
\multicolumn{1}{|r|}{}                           & $-$   & \multicolumn{1}{r|}{306}  & \multicolumn{1}{r|}{2,607} & \multicolumn{1}{r|}{2,913} & \multicolumn{1}{r|}{72\%}  \\ \cline{2-6} 
\multicolumn{1}{|r|}{}                           & Total & \multicolumn{1}{r|}{652}  & \multicolumn{1}{r|}{3,368} & \multicolumn{1}{r|}{4,020} & \multicolumn{1}{r|}{100\%} \\ \cline{2-6} 
\multicolumn{1}{|r|}{}                           & \%    & \multicolumn{1}{r|}{16\%} & \multicolumn{1}{r|}{84\%}  & \multicolumn{1}{r|}{100\%} &                            \\ \cline{1-5}
\end{tabular}
\label{tab:conf}
\end{table}
% Overall confusion
%%%%%%%%%%%%%%%%%%%%%%%%%%%%%%%%%%%%%%%%%%%%%%%%%%%%

Table \ref{tab:conf} presents the overall confusion matrix. Among the 4,020 cases, 1,107 are forecasted to be recidivists, while only 652 actually recidivate within the next two years. The overall accuracy of the system is 73\%. Among the 1,107 who are predicted as risky, only 346 recidivate hence a precision of 31\%. Among the 652 who actually recidivate, only 346 were predicted as risky hence a recall of 53\%. Since larger values indicate better performance for these metrics, the results indicate a questionable performance, especially considering the harms that may result from potential misjudgments in the criminal justice system where COMPAS is utilized in.

Next, fairness studies are performed, and findings are reported for sex, race, and intersectional groups.

%%%%%%%%%%%%%%%%%%%%%%%%%%%%%%%%%%%%%%%%%%%%%%%%%%%%
\begin{table*}[t]
\caption{Contingency matrix based on sex: observed (O), expected (E), and adjusted standardized residual (R).}

\centering
\begin{tabular}{|r|crrcrr|crrcrr|}
\hline
Actual    & \multicolumn{6}{c|}{$+$}                                                                                                                                       & \multicolumn{6}{c|}{$-$}                                                                                                                                          \\ \hline
Predicted & \multicolumn{3}{c|}{$+$}                                                        & \multicolumn{3}{c|}{$-$}                                                     & \multicolumn{3}{c|}{$+$}                                                        & \multicolumn{3}{c|}{$-$}                                                        \\ \hline
O/E/R     & \multicolumn{1}{c|}{O}   & \multicolumn{1}{c|}{E}   & \multicolumn{1}{c|}{R}    & \multicolumn{1}{c|}{O}   & \multicolumn{1}{c|}{E}   & \multicolumn{1}{c|}{R} & \multicolumn{1}{c|}{O}   & \multicolumn{1}{c|}{E}   & \multicolumn{1}{c|}{R}    & \multicolumn{1}{c|}{O}    & \multicolumn{1}{c|}{E}     & \multicolumn{1}{c|}{R} \\ \hline
Female    & \multicolumn{1}{r|}{27}  & \multicolumn{1}{r|}{72}  & \multicolumn{1}{r|}{\textbf{-6.3}} & \multicolumn{1}{r|}{50}  & \multicolumn{1}{r|}{64}  & -2.0                   & \multicolumn{1}{r|}{137} & \multicolumn{1}{r|}{159} & \multicolumn{1}{r|}{-2.2} & \multicolumn{1}{r|}{627}  & \multicolumn{1}{r|}{545}   & \textbf{6.6 }                   \\ \hline
Male      & \multicolumn{1}{r|}{319} & \multicolumn{1}{r|}{274} & \multicolumn{1}{r|}{\textbf{6.3}}  & \multicolumn{1}{r|}{256} & \multicolumn{1}{r|}{242} & 2.0                    & \multicolumn{1}{r|}{624} & \multicolumn{1}{r|}{602} & \multicolumn{1}{r|}{2.2}  & \multicolumn{1}{r|}{1980} & \multicolumn{1}{r|}{2,062} & -\textbf{6.6}                   \\ \hline
\end{tabular}
\label{tab:sex_contingency}
\end{table*}
% Sex Contingency
%%%%%%%%%%%%%%%%%%%%%%%%%%%%%%%%%%%%%%%%%%%%%%%%%%%%

%%%%%%%%%%%%%%%%%%%%%%%%%%%%%%%%%%%%%%%%%%%%%%%%%%%%
\begin{table*}[t]
\caption{Contingency matrix based on race: observed (O), expected (E), and adjusted standardized residual (R).}

\centering
\begin{tabular}{|r|rrrrrr|rrrrrr|}
\hline
Actual           & \multicolumn{6}{c|}{$+$}                                                                                                                                       & \multicolumn{6}{c|}{$-$}                                                                                                                                           \\ \hline
Predicted        & \multicolumn{3}{c|}{$+$}                                                        & \multicolumn{3}{c|}{$-$}                                                     & \multicolumn{3}{c|}{$+$}                                                        & \multicolumn{3}{c|}{$-$}                                                         \\ \hline
O/E/R            & \multicolumn{1}{c|}{O}   & \multicolumn{1}{c|}{E}   & \multicolumn{1}{c|}{R}    & \multicolumn{1}{c|}{O}   & \multicolumn{1}{c|}{E}   & \multicolumn{1}{c|}{R} & \multicolumn{1}{c|}{O}   & \multicolumn{1}{c|}{E}   & \multicolumn{1}{c|}{R}    & \multicolumn{1}{c|}{O}     & \multicolumn{1}{c|}{E}     & \multicolumn{1}{c|}{R} \\ \hline
African-American & \multicolumn{1}{r|}{250} & \multicolumn{1}{r|}{165} & \multicolumn{1}{r|}{\textbf{9.6}}  & \multicolumn{1}{r|}{154} & \multicolumn{1}{r|}{146} & 1.0                    & \multicolumn{1}{r|}{468} & \multicolumn{1}{r|}{363} & \multicolumn{1}{r|}{\textbf{8.5}}  & \multicolumn{1}{r|}{1,046} & \multicolumn{1}{r|}{1,244} & \textbf{-13.1}                  \\ \hline
Asian            & \multicolumn{1}{r|}{3}   & \multicolumn{1}{r|}{2}   & \multicolumn{1}{r|}{0.5}  & \multicolumn{1}{r|}{0}   & \multicolumn{1}{r|}{2}   & -1.5                   & \multicolumn{1}{r|}{1}   & \multicolumn{1}{r|}{5}   & \multicolumn{1}{r|}{-2.0} & \multicolumn{1}{r|}{22}    & \multicolumn{1}{r|}{17}    & 2.1                    \\ \hline
Caucasian        & \multicolumn{1}{r|}{64}  & \multicolumn{1}{r|}{126} & \multicolumn{1}{r|}{\textbf{-7.2}} & \multicolumn{1}{r|}{110} & \multicolumn{1}{r|}{111} & -0.1                   & \multicolumn{1}{r|}{198} & \multicolumn{1}{r|}{276} & \multicolumn{1}{r|}{\textbf{-6.5}} & \multicolumn{1}{r|}{1,087} & \multicolumn{1}{r|}{946}   & \textbf{9.7}                   \\ \hline
Hispanic         & \multicolumn{1}{r|}{10}  & \multicolumn{1}{r|}{31}  & \multicolumn{1}{r|}{\textbf{-4.1}} & \multicolumn{1}{r|}{25}  & \multicolumn{1}{r|}{27}  & -0.4                   & \multicolumn{1}{r|}{61}  & \multicolumn{1}{r|}{67}  & \multicolumn{1}{r|}{-0.9} & \multicolumn{1}{r|}{259}   & \multicolumn{1}{r|}{230}   & \textbf{3.4}                    \\ \hline
Native American  & \multicolumn{1}{r|}{1}   & \multicolumn{1}{r|}{1}   & \multicolumn{1}{r|}{0.5}  & \multicolumn{1}{r|}{0}   & \multicolumn{1}{r|}{1}   & -0.8                   & \multicolumn{1}{r|}{1}   & \multicolumn{1}{r|}{1}   & \multicolumn{1}{r|}{-0.3} & \multicolumn{1}{r|}{5}     & \multicolumn{1}{r|}{5}     & 0.4                    \\ \hline
Other            & \multicolumn{1}{r|}{18}  & \multicolumn{1}{r|}{22}  & \multicolumn{1}{r|}{-0.9} & \multicolumn{1}{r|}{17}  & \multicolumn{1}{r|}{19}  & -0.6                   & \multicolumn{1}{r|}{32}  & \multicolumn{1}{r|}{48}  & \multicolumn{1}{r|}{-2.7} & \multicolumn{1}{r|}{188}   & \multicolumn{1}{r|}{165}   & 3.1                    \\ \hline
\end{tabular}
\label{tab:race_contingency}
\end{table*}
% Race Contingency
%%%%%%%%%%%%%%%%%%%%%%%%%%%%%%%%%%%%%%%%%%%%%%%%%%%%

\begin{table}[t]

\end{table}

%%%%%%%%%%%%%%%%%%%%%%%%%%%%%%%%%%%%%%%%%%%%%%%%%%%%%%%%%%%%%%%%%%
\begin{table}[htbp]
\caption{Distribution of the confusion matrix based on sex. (T) A view with predictions as the basis. (B) A view with actual (ground truth) values as the basis.}

\setlength\tabcolsep{5pt}
\centering

\label{tab:criteria}
\begin{tabular}{|r|r|r|r|r|r|r|} 
\hline
Predicted & \multicolumn{3}{c|}{+}                                                                                                                                                                   & \multicolumn{3}{c|}{-}                                                                                                                                                                   \\ 
\hline
Actual    & \multicolumn{1}{c|}{+}                                                  & \multicolumn{1}{c|}{-}                                & \multicolumn{1}{c|}{Total}                             & \multicolumn{1}{c|}{+}                               & \multicolumn{1}{c|}{-}                                                  & \multicolumn{1}{c|}{Total}                              \\ 
\hline
Female    & \begin{tabular}[c]{@{}r@{}}\textbf{3\% }\\\textbf{ (16\%)}\end{tabular} & \begin{tabular}[c]{@{}r@{}}17\%\\ (84\%)\end{tabular} & \begin{tabular}[c]{@{}r@{}}19\%\\ (100\%)\end{tabular} & \begin{tabular}[c]{@{}r@{}}6\%\\ (7\%)\end{tabular}  & \begin{tabular}[c]{@{}r@{}}\textbf{75\%}\\\textbf{ (93\%)}\end{tabular} & \begin{tabular}[c]{@{}r@{}}81\%\\ (100\%)\end{tabular}  \\ 
\hline
Male      & \begin{tabular}[c]{@{}r@{}}\textbf{10\%}\\\textbf{ (34\%)}\end{tabular} & \begin{tabular}[c]{@{}r@{}}20\%\\ (66\%)\end{tabular} & \begin{tabular}[c]{@{}r@{}}30\%\\ (100\%)\end{tabular} & \begin{tabular}[c]{@{}r@{}}8\%\\ (11\%)\end{tabular} & \begin{tabular}[c]{@{}r@{}}\textbf{62\%}\\\textbf{ (89\%)}\end{tabular} & \begin{tabular}[c]{@{}r@{}}70\%\\ (100\%)\end{tabular}  \\
\hline
\end{tabular}

\vspace{1em}
\begin{tabular}{|r|r|r|r|r|r|r|} 
\hline
Actual    & \multicolumn{3}{c|}{+}                                                                                                                                                                     & \multicolumn{3}{c|}{-}                                                                                                                                                                       \\ 
\hline
Predicted & \multicolumn{1}{c|}{+}                                                   & \multicolumn{1}{c|}{-}                                & \multicolumn{1}{c|}{Total}                              & \multicolumn{1}{c|}{+}                                 & \multicolumn{1}{c|}{-}                                                   & \multicolumn{1}{c|}{Total}                               \\ 
\hline
Female    & \begin{tabular}[c]{@{}r@{}}\textbf{3\%}\\\textbf{ (35\%)}\end{tabular}   & \begin{tabular}[c]{@{}r@{}}6\% \\ (65\%)\end{tabular} & \begin{tabular}[c]{@{}r@{}}9\% \\ (100\%)\end{tabular}  & \begin{tabular}[c]{@{}r@{}}16\% \\ (18\%)\end{tabular} & \begin{tabular}[c]{@{}r@{}}\textbf{75\% }\\\textbf{ (82\%)}\end{tabular} & \begin{tabular}[c]{@{}r@{}}91\% \\ (100\%)\end{tabular}  \\ 
\hline
Male      & \begin{tabular}[c]{@{}r@{}}\textbf{10\% }\\\textbf{ (55\%)}\end{tabular} & \begin{tabular}[c]{@{}r@{}}8\% \\ (45\%)\end{tabular} & \begin{tabular}[c]{@{}r@{}}18\% \\ (100\%)\end{tabular} & \begin{tabular}[c]{@{}r@{}}20\% \\ (24\%)\end{tabular} & \begin{tabular}[c]{@{}r@{}}\textbf{62\% }\\\textbf{ (76\%)}\end{tabular} & \begin{tabular}[c]{@{}r@{}}82\% \\ (100\%)\end{tabular}  \\
\hline
\end{tabular}
\label{tab:sex_ratio}
\end{table}
% Sex Confusion
%%%%%%%%%%%%%%%%%%%%%%%%%%%%%%%%%%%%%%%%%%%%%%%%%%%%%%%%%%%%%%%%%%

%%%%%%%%%%%%%%%%%%%%%%%%%%%%%%%%%%%%%%%%%%%%%%%%%%%%%%%%%%%%%%%%%%
\begin{table}[htbp]
\caption{Distribution of the confusion matrix based on race. (T) A view with predictions as the basis. (B) A view with actual (ground truth) values as the basis.}
\setlength\tabcolsep{5pt}
\centering
\begin{tabular}{|r|rrr|rrr|}
\hline
Predicted                                                   & \multicolumn{3}{c|}{+}                                                                                                                                                                                                                & \multicolumn{3}{c|}{-}                                                                                                                                                                                                      \\ \hline
Actual                                                      & \multicolumn{1}{c|}{+}                                                               & \multicolumn{1}{c|}{-}                                                               & \multicolumn{1}{c|}{Total}                              & \multicolumn{1}{c|}{+}                                                     & \multicolumn{1}{c|}{-}                                                               & \multicolumn{1}{c|}{Total}                              \\ \hline
\begin{tabular}[c]{@{}r@{}}African-\\ American\end{tabular} & \multicolumn{1}{r|}{\textbf{\begin{tabular}[c]{@{}r@{}}13\% \\ (35\%)\end{tabular}}} & \multicolumn{1}{r|}{\textbf{\begin{tabular}[c]{@{}r@{}}24\% \\ (65\%)\end{tabular}}} & \begin{tabular}[c]{@{}r@{}}37\% \\ (100\%)\end{tabular} & \multicolumn{1}{r|}{\begin{tabular}[c]{@{}r@{}}8\% \\ (13\%)\end{tabular}} & \multicolumn{1}{r|}{\textbf{\begin{tabular}[c]{@{}r@{}}55\% \\ (87\%)\end{tabular}}} & \begin{tabular}[c]{@{}r@{}}63\% \\ (100\%)\end{tabular} \\ \hline
Asian                                                       & \multicolumn{1}{r|}{\begin{tabular}[c]{@{}r@{}}12\% \\ (75\%)\end{tabular}}          & \multicolumn{1}{r|}{\begin{tabular}[c]{@{}r@{}}4\% \\ (25\%)\end{tabular}}           & \begin{tabular}[c]{@{}r@{}}15\% \\ (100\%)\end{tabular} & \multicolumn{1}{r|}{\begin{tabular}[c]{@{}r@{}}0\% \\ (0\%)\end{tabular}}  & \multicolumn{1}{r|}{\begin{tabular}[c]{@{}r@{}}85\% \\ (100\%)\end{tabular}}         & \begin{tabular}[c]{@{}r@{}}85\% \\ (100\%)\end{tabular} \\ \hline
Caucasian                                                   & \multicolumn{1}{r|}{\textbf{\begin{tabular}[c]{@{}r@{}}4\% \\ (24\%)\end{tabular}}}  & \multicolumn{1}{r|}{\textbf{\begin{tabular}[c]{@{}r@{}}14\% \\ (76\%)\end{tabular}}} & \begin{tabular}[c]{@{}r@{}}18\% \\ (100\%)\end{tabular} & \multicolumn{1}{r|}{\begin{tabular}[c]{@{}r@{}}8\% \\ (9\%)\end{tabular}}  & \multicolumn{1}{r|}{\textbf{\begin{tabular}[c]{@{}r@{}}75\% \\ (91\%)\end{tabular}}} & \begin{tabular}[c]{@{}r@{}}82\% \\ (100\%)\end{tabular} \\ \hline
Hispanic                                                    & \multicolumn{1}{r|}{\textbf{\begin{tabular}[c]{@{}r@{}}3\% \\ (14\%)\end{tabular}}}  & \multicolumn{1}{r|}{\begin{tabular}[c]{@{}r@{}}17\% \\ (86\%)\end{tabular}}          & \begin{tabular}[c]{@{}r@{}}20\% \\ (100\%)\end{tabular} & \multicolumn{1}{r|}{\begin{tabular}[c]{@{}r@{}}7\% \\ (9\%)\end{tabular}}  & \multicolumn{1}{r|}{\textbf{\begin{tabular}[c]{@{}r@{}}73\% \\ (91\%)\end{tabular}}} & \begin{tabular}[c]{@{}r@{}}80\% \\ (100\%)\end{tabular} \\ \hline
\begin{tabular}[c]{@{}r@{}}Native \\ American\end{tabular}  & \multicolumn{1}{r|}{\begin{tabular}[c]{@{}r@{}}14\% \\ (50\%)\end{tabular}}          & \multicolumn{1}{r|}{\begin{tabular}[c]{@{}r@{}}14\% \\ (50\%)\end{tabular}}          & \begin{tabular}[c]{@{}r@{}}29\% \\ (100\%)\end{tabular} & \multicolumn{1}{r|}{\begin{tabular}[c]{@{}r@{}}0\% \\ (0\%)\end{tabular}}  & \multicolumn{1}{r|}{\begin{tabular}[c]{@{}r@{}}71\% \\ (100\%)\end{tabular}}         & \begin{tabular}[c]{@{}r@{}}71\% \\ (100\%)\end{tabular} \\ \hline
Other                                                       & \multicolumn{1}{r|}{\begin{tabular}[c]{@{}r@{}}7\% \\ (36\%)\end{tabular}}           & \multicolumn{1}{r|}{\begin{tabular}[c]{@{}r@{}}13\% \\ (64\%)\end{tabular}}          & \begin{tabular}[c]{@{}r@{}}20\% \\ (100\%)\end{tabular} & \multicolumn{1}{r|}{\begin{tabular}[c]{@{}r@{}}7\% \\ (8\%)\end{tabular}}  & \multicolumn{1}{r|}{\begin{tabular}[c]{@{}r@{}}74\% \\ (92\%)\end{tabular}}          & \begin{tabular}[c]{@{}r@{}}80\% \\ (100\%)\end{tabular} \\ \hline
\end{tabular}

\vspace{1em}

\begin{tabular}{|r|rrr|rrr|}
\hline
Actual                                                      & \multicolumn{3}{c|}{+}                                                                                                                                                                                                      & \multicolumn{3}{c|}{-}                                                                                                                                                                                                                \\ \hline
Predicted                                                   & \multicolumn{1}{c|}{+}                                                               & \multicolumn{1}{c|}{-}                                                     & \multicolumn{1}{c|}{Total}                              & \multicolumn{1}{c|}{+}                                                               & \multicolumn{1}{c|}{-}                                                               & \multicolumn{1}{c|}{Total}                              \\ \hline
\begin{tabular}[c]{@{}r@{}}African-\\ American\end{tabular} & \multicolumn{1}{r|}{\textbf{\begin{tabular}[c]{@{}r@{}}13\% \\ (62\%)\end{tabular}}} & \multicolumn{1}{r|}{\begin{tabular}[c]{@{}r@{}}8\% \\ (38\%)\end{tabular}} & \begin{tabular}[c]{@{}r@{}}21\% \\ (100\%)\end{tabular} & \multicolumn{1}{r|}{\textbf{\begin{tabular}[c]{@{}r@{}}24\% \\ (31\%)\end{tabular}}} & \multicolumn{1}{r|}{\textbf{\begin{tabular}[c]{@{}r@{}}55\% \\ (69\%)\end{tabular}}} & \begin{tabular}[c]{@{}r@{}}79\% \\ (100\%)\end{tabular} \\ \hline
Asian                                                       & \multicolumn{1}{r|}{\begin{tabular}[c]{@{}r@{}}12\% \\ (100\%)\end{tabular}}         & \multicolumn{1}{r|}{\begin{tabular}[c]{@{}r@{}}0\% \\ (0\%)\end{tabular}}  & \begin{tabular}[c]{@{}r@{}}12\% \\ (100\%)\end{tabular} & \multicolumn{1}{r|}{\begin{tabular}[c]{@{}r@{}}4\% \\ (4\%)\end{tabular}}            & \multicolumn{1}{r|}{\begin{tabular}[c]{@{}r@{}}85\% \\ (96\%)\end{tabular}}          & \begin{tabular}[c]{@{}r@{}}88\% \\ (100\%)\end{tabular} \\ \hline
Caucasian                                                   & \multicolumn{1}{r|}{\textbf{\begin{tabular}[c]{@{}r@{}}4\% \\ (37\%)\end{tabular}}}  & \multicolumn{1}{r|}{\begin{tabular}[c]{@{}r@{}}8\% \\ (63\%)\end{tabular}} & \begin{tabular}[c]{@{}r@{}}12\% \\ (100\%)\end{tabular} & \multicolumn{1}{r|}{\textbf{\begin{tabular}[c]{@{}r@{}}14\% \\ (15\%)\end{tabular}}} & \multicolumn{1}{r|}{\textbf{\begin{tabular}[c]{@{}r@{}}75\% \\ (85\%)\end{tabular}}} & \begin{tabular}[c]{@{}r@{}}88\% \\ (100\%)\end{tabular} \\ \hline
Hispanic                                                    & \multicolumn{1}{r|}{\textbf{\begin{tabular}[c]{@{}r@{}}3\% \\ (29\%)\end{tabular}}}  & \multicolumn{1}{r|}{\begin{tabular}[c]{@{}r@{}}7\% \\ (71\%)\end{tabular}} & \begin{tabular}[c]{@{}r@{}}10\% \\ (100\%)\end{tabular} & \multicolumn{1}{r|}{\begin{tabular}[c]{@{}r@{}}17\% \\ (19\%)\end{tabular}}          & \multicolumn{1}{r|}{\textbf{\begin{tabular}[c]{@{}r@{}}73\% \\ (81\%)\end{tabular}}} & \begin{tabular}[c]{@{}r@{}}90\% \\ (100\%)\end{tabular} \\ \hline
\begin{tabular}[c]{@{}r@{}}Native \\ American\end{tabular}  & \multicolumn{1}{r|}{\begin{tabular}[c]{@{}r@{}}14\% \\ (100\%)\end{tabular}}         & \multicolumn{1}{r|}{\begin{tabular}[c]{@{}r@{}}0\% \\ (0\%)\end{tabular}}  & \begin{tabular}[c]{@{}r@{}}14\% \\ (100\%)\end{tabular} & \multicolumn{1}{r|}{\begin{tabular}[c]{@{}r@{}}14\% \\ (17\%)\end{tabular}}          & \multicolumn{1}{r|}{\begin{tabular}[c]{@{}r@{}}71\% \\ (83\%)\end{tabular}}          & \begin{tabular}[c]{@{}r@{}}86\% \\ (100\%)\end{tabular} \\ \hline
Other                                                       & \multicolumn{1}{r|}{\begin{tabular}[c]{@{}r@{}}7\% \\ (51\%)\end{tabular}}           & \multicolumn{1}{r|}{\begin{tabular}[c]{@{}r@{}}7\% \\ (49\%)\end{tabular}} & \begin{tabular}[c]{@{}r@{}}14\% \\ (100\%)\end{tabular} & \multicolumn{1}{r|}{\begin{tabular}[c]{@{}r@{}}13\% \\ (15\%)\end{tabular}}          & \multicolumn{1}{r|}{\begin{tabular}[c]{@{}r@{}}74\% \\ (85\%)\end{tabular}}          & \begin{tabular}[c]{@{}r@{}}86\% \\ (100\%)\end{tabular} \\ \hline
\end{tabular}

\label{tab:race_ratio}
\end{table}
% Race Confusion
%%%%%%%%%%%%%%%%%%%%%%%%%%%%%%%%%%%%%%%%%%%%%%%%%%%%%%%%%%%%%%%%%%

\subsubsection{Sex}

The equal confusion test resulted in $p < 0.001$, which indicated a statistically significant association between sex and cells of the confusion matrix. As this indicates that the system is unfair, the confusion parity error is computed as $\phi = 0.12$. According to the corresponding interpretation presented in Table \ref{tab:phinterpretation} with $q=2, r=4$, it can be concluded that the system exhibits small but statistically significant unfairness. A post hoc analysis follows this finding to identify the specific sources of unfairness.

Table \ref{tab:sex_contingency} presents the observed values (O), expected values (\textit{E}), and adjusted standardized residuals (R) for the contingency matrix. Assuming a desired $p < 0.001$, the absolute two-tailed critical value is $3.29$. The significant values are shown in boldface type. To further investigate the characteristics of the unfairness and impacted groups, Table \ref{tab:sex_ratio} presents confusion matrices based on sex. The cell values are presented as the proportion of row totals, and as the proportion of subtotals in the case of parenthesized values. The cells corresponding to the significant values are shown in boldface type. A closer analysis of the significant cells results in the following observations.

\begin{itemize}
    \item Among predicted risky females, only $16\%$ are actually recidivists compared to the same figure of $34\%$ for males. Precision is higher for males than females. Hence, females are more likely to be incorrectly predicted as risky.
    \item Among predicted non-risky females, $93\%$ are actual non-recidivists, while the same figure goes down to $89\%$ for males. Negative predictive value is higher for females than males. Hence, males are more likely to benefit from false negatives.
    \item Among actual recidivist females, only $35\%$ are correctly predicted as risky compared to the same figure of $55\%$ for males. Recall is higher for males than females. Hence, females are more likely to benefit from under-identification of risky status.
    \item Among actual non-recidivist females, $82\%$ are correctly predicted as non-risky, while the same figure goes down to $76\%$ for males. Specificity is higher for females than males. Hence, males are more likely to suffer from an under-identification of non-risky status.
\end{itemize}

The first two findings reveal a disadvantageous position for females, whereas the last two findings indicate an advantageous position, compared to males. These findings are not contradictory as they are based on different measurements. It indicates that implications from a fairness analysis are not straightforward and require a comprehensive perspective rather than an inspection of a subset of measurements.

\subsubsection{Race}

The same analysis is repeated for race. The equal confusion test resulted in $p < 0.001$. Subsequently, the confusion parity error is computed as $\phi = 0.13$, which indicates a small but statistically significant unfairness with $q=6, r=4$. Tables \ref{tab:race_contingency} and \ref{tab:race_ratio} presents the contingency and confusion matrices in the same fashion as Tables \ref{tab:sex_contingency} and \ref{tab:sex_ratio}. A closer analysis of the significant cells results in the following observations.

\begin{itemize}
    \item Among predicted risky African-Americans, $35\%$  are actually recidivists compared to  $24\%$ and $14\%$ for Caucasians and Hispanics, respectively. Precision is highest for African-Americans and lowest for Hispanics. Hence, Hispanics are more likely to be incorrectly predicted as risky than Caucasians and African-Americans.
    
    \item Among predicted non-risky Caucasians and Hispanics, $91\%$ are actual non-recidivists, while the same figure goes down to $87\%$ for African-Americans. Negative predictive value is lower for African-Americans. Hence, African-Americans are more likely to benefit from false negatives.
    
    \item Among actual recidivist Caucasians and Hispanics, only $37\%$ and $29\%$, respectively, are correctly predicted as risky compared to the same figure of $62\%$ for African-Americans. Recall is higher for African-Americans, Caucasians, and Hispanics are more likely to benefit from under-identification of risky status.
    
    \item Among actual non-recidivist African-Americans, $69\%$ are incorrectly predicted as risky compared to $85\%$ and $81\%$ for Caucasians and Hispanics, respectively. Specificity is lower for African-Americans. Hence, African-Americans are more likely to suffer from an under-identification of non-risky status.
\end{itemize}

The first two findings indicate an advantageous position for African-Americans from one perspective, whereas the last two indicate a disadvantageous one from another.

\subsubsection{Intersectional Groups}

%%%%%%%%%%%%%%%%%%%%%%%%%%%%%%%%%%%%%%%%%%%%%%%%%%%%
\begin{table*}[t]
\caption{Contingency matrix based on intersectional groups: observed (O), expected (E), and adjusted standardized residual (R).}
\centering
\begin{tabular}{|rr|rrrrrr|rrrrrr|}
\hline
\multicolumn{2}{|r|}{Actual}                                     & \multicolumn{6}{c|}{$+$}                                                                                                                                                                    & \multicolumn{6}{c|}{$-$}                                                                                                                                                           \\ \hline
\multicolumn{2}{|r|}{Predicted}                                  & \multicolumn{3}{c|}{$+$}                                                                            & \multicolumn{3}{c|}{$-$}                                                              & \multicolumn{3}{c|}{$+$}                                                                   & \multicolumn{3}{c|}{$-$}                                                              \\ \hline
\multicolumn{2}{|r|}{O/E/R}                                      & \multicolumn{1}{c|}{O}   & \multicolumn{1}{c|}{E}            & \multicolumn{1}{c|}{R}               & \multicolumn{1}{c|}{O}   & \multicolumn{1}{c|}{E}            & \multicolumn{1}{c|}{R} & \multicolumn{1}{c|}{O}   & \multicolumn{1}{c|}{E}   & \multicolumn{1}{c|}{R}               & \multicolumn{1}{c|}{O}   & \multicolumn{1}{c|}{E}            & \multicolumn{1}{c|}{R} \\ \hline
\multicolumn{1}{|r|}{\multirow{5}{*}{Female}} & African-American & \multicolumn{1}{r|}{19}  & \multicolumn{1}{r|}{\textit{34}}  & \multicolumn{1}{r|}{(-2.8)}          & \multicolumn{1}{r|}{28}  & \multicolumn{1}{r|}{\textit{30}}  & (-0.4)                 & \multicolumn{1}{r|}{81}  & \multicolumn{1}{r|}{74}  & \multicolumn{1}{r|}{(0.9)}           & \multicolumn{1}{r|}{265} & \multicolumn{1}{r|}{\textit{255}} & (1.1)                  \\ \cline{2-14} 
\multicolumn{1}{|r|}{}                        & Asian            & \multicolumn{1}{r|}{0}   & \multicolumn{1}{r|}{\textit{0}}   & \multicolumn{1}{r|}{(-0.3)}          & \multicolumn{1}{r|}{0}   & \multicolumn{1}{r|}{\textit{0}}   & (-0.3)                 & \multicolumn{1}{r|}{0}   & \multicolumn{1}{r|}{0}   & \multicolumn{1}{r|}{(-0.5)}          & \multicolumn{1}{r|}{1}   & \multicolumn{1}{r|}{\textit{1}}   & (0.7)                  \\ \cline{2-14} 
\multicolumn{1}{|r|}{}                        & Caucasian        & \multicolumn{1}{r|}{8}   & \multicolumn{1}{r|}{\textit{29}}  & \multicolumn{1}{r|}{\textbf{(-4.3)}} & \multicolumn{1}{r|}{15}  & \multicolumn{1}{r|}{\textit{26}}  & (-2.3)                 & \multicolumn{1}{r|}{41}  & \multicolumn{1}{r|}{64}  & \multicolumn{1}{r|}{\textbf{(-3.3)}} & \multicolumn{1}{r|}{272} & \multicolumn{1}{r|}{\textit{218}} & \textbf{(6.5)}         \\ \cline{2-14} 
\multicolumn{1}{|r|}{}                        & Hispanic         & \multicolumn{1}{r|}{0}   & \multicolumn{1}{r|}{\textit{5}}   & \multicolumn{1}{r|}{(-2.4)}          & \multicolumn{1}{r|}{5}   & \multicolumn{1}{r|}{\textit{5}}   & (0.2)                  & \multicolumn{1}{r|}{5}   & \multicolumn{1}{r|}{12}  & \multicolumn{1}{r|}{(-2.2)}          & \multicolumn{1}{r|}{51}  & \multicolumn{1}{r|}{\textit{40}}  & (3.1)                  \\ \cline{2-14} 
\multicolumn{1}{|r|}{}                        & Other            & \multicolumn{1}{r|}{0}   & \multicolumn{1}{r|}{\textit{4}}   & \multicolumn{1}{r|}{(-2.2)}          & \multicolumn{1}{r|}{2}   & \multicolumn{1}{r|}{\textit{4}}   & (-1.0)                 & \multicolumn{1}{r|}{10}  & \multicolumn{1}{r|}{9}   & \multicolumn{1}{r|}{(0.2)}           & \multicolumn{1}{r|}{38}  & \multicolumn{1}{r|}{\textit{32}}  & (1.7)                  \\ \hline
\multicolumn{1}{|r|}{\multirow{6}{*}{Male}}   & African-American & \multicolumn{1}{r|}{231} & \multicolumn{1}{r|}{\textit{131}} & \multicolumn{1}{r|}{\textbf{(11.6)}} & \multicolumn{1}{r|}{126} & \multicolumn{1}{r|}{\textit{116}} & (1.2)                  & \multicolumn{1}{r|}{387} & \multicolumn{1}{r|}{289} & \multicolumn{1}{r|}{\textbf{(8.2)}}  & \multicolumn{1}{r|}{781} & \multicolumn{1}{r|}{\textit{989}} & \textbf{(-14.2)}       \\ \cline{2-14} 
\multicolumn{1}{|r|}{}                        & Asian            & \multicolumn{1}{r|}{3}   & \multicolumn{1}{r|}{2}            & \multicolumn{1}{r|}{(0.6)}           & \multicolumn{1}{r|}{0}   & \multicolumn{1}{r|}{2}            & (-1.4)                 & \multicolumn{1}{r|}{1}   & \multicolumn{1}{r|}{5}   & \multicolumn{1}{r|}{(-1.9)}          & \multicolumn{1}{r|}{21}  & \multicolumn{1}{r|}{16}           & (2.0)                  \\ \cline{2-14} 
\multicolumn{1}{|r|}{}                        & Caucasian        & \multicolumn{1}{r|}{56}  & \multicolumn{1}{r|}{97}           & \multicolumn{1}{r|}{\textbf{(-5.1)}} & \multicolumn{1}{r|}{95}  & \multicolumn{1}{r|}{85}           & (1.3)                  & \multicolumn{1}{r|}{157} & \multicolumn{1}{r|}{213} & \multicolumn{1}{r|}{\textbf{(-5.0)}} & \multicolumn{1}{r|}{815} & \multicolumn{1}{r|}{728}          & \textbf{(6.4)}         \\ \cline{2-14} 
\multicolumn{1}{|r|}{}                        & Hispanic         & \multicolumn{1}{r|}{10}  & \multicolumn{1}{r|}{25}           & \multicolumn{1}{r|}{\textbf{(-3.3)}} & \multicolumn{1}{r|}{20}  & \multicolumn{1}{r|}{22}           & (-0.5)                 & \multicolumn{1}{r|}{56}  & \multicolumn{1}{r|}{56}  & \multicolumn{1}{r|}{(0.1)}           & \multicolumn{1}{r|}{208} & \multicolumn{1}{r|}{191}          & (2.2)                  \\ \cline{2-14} 
\multicolumn{1}{|r|}{}                        & Native American  & \multicolumn{1}{r|}{1}   & \multicolumn{1}{r|}{1}            & \multicolumn{1}{r|}{(0.5)}           & \multicolumn{1}{r|}{0}   & \multicolumn{1}{r|}{1}            & (-0.8)                 & \multicolumn{1}{r|}{1}   & \multicolumn{1}{r|}{1}   & \multicolumn{1}{r|}{(-0.3)}          & \multicolumn{1}{r|}{1}   & \multicolumn{1}{r|}{5}            & (0.4)                  \\ \cline{2-14} 
\multicolumn{1}{|r|}{}                        & Other            & \multicolumn{1}{r|}{18}  & \multicolumn{1}{r|}{18}           & \multicolumn{1}{r|}{(0.1)}           & \multicolumn{1}{r|}{15}  & \multicolumn{1}{r|}{16}           & (-0.2)                 & \multicolumn{1}{r|}{22}  & \multicolumn{1}{r|}{39}  & \multicolumn{1}{r|}{(-3.1)}          & \multicolumn{1}{r|}{18}  & \multicolumn{1}{r|}{133}          & (2.6)                  \\ \hline
\end{tabular}
\label{tab:is_contingency}
\end{table*}
%Intersection Contingency
%%%%%%%%%%%%%%%%%%%%%%%%%%%%%%%%%%%%%%%%%%%%%%%%%%%%

%%%%%%%%%%%%%%%%%%%%%%%%%%%%%%%%%%%%%%%%%%%%%%%%%%%%%%%%%%%%%%%%%%
\begin{table}[t]
\caption{Distribution of the confusion matrix based on intersectional groups. (T) A view with predictions as the basis. (B) A view with actual (ground truth) values as the basis.}
\setlength\tabcolsep{3pt}
\centering
\begin{tabular}{|r|r|r|r|r|r|r|r|} 
\cline{2-8}
\multicolumn{1}{r|}{}   & Predicted                                                   & \multicolumn{3}{c|}{+}                                                                                                                                                                                        & \multicolumn{3}{c|}{-}                                                                                                                                                                     \\ 
\cline{2-8}
\multicolumn{1}{r|}{}   & Actual                                                      & \multicolumn{1}{c|}{+}                                                   & \multicolumn{1}{c|}{-}                                                   & \multicolumn{1}{c|}{Total}                              & \multicolumn{1}{c|}{+}                               & \multicolumn{1}{c|}{-}                                                   & \multicolumn{1}{c|}{Total}                               \\ 
\hline
\multirow{5}{*}{Female} & \begin{tabular}[c]{@{}r@{}}African-\\ American\end{tabular} & \begin{tabular}[c]{@{}r@{}}5\%\\ (19\%)\end{tabular}                     & \begin{tabular}[c]{@{}r@{}}21\% \\ (81\%)\end{tabular}                   & \begin{tabular}[c]{@{}r@{}}25\% \\ (100\%)\end{tabular} & \begin{tabular}[c]{@{}r@{}}7\%\\ (10\%)\end{tabular} & \begin{tabular}[c]{@{}r@{}}67\%\\ (90\%)\end{tabular}                    & \begin{tabular}[c]{@{}r@{}}75\% \\ (100\%)\end{tabular}  \\ 
\cline{2-8}
                        & Asian                                                       & \begin{tabular}[c]{@{}r@{}}0\%\\ (n/a)\end{tabular}                      & \begin{tabular}[c]{@{}r@{}}0\%\\ (n/a)\end{tabular}                      & \begin{tabular}[c]{@{}r@{}}0\% \\ (100\%)\end{tabular}  & \begin{tabular}[c]{@{}r@{}}0\%\\ (0\%)\end{tabular}  & \begin{tabular}[c]{@{}r@{}}100\% \\ (100\%)\end{tabular}                 & \begin{tabular}[c]{@{}r@{}}100\%\\ (100\%)\end{tabular}  \\ 
\cline{2-8}
                        & Caucasian                                                   & \begin{tabular}[c]{@{}r@{}}\textbf{2\% }\\\textbf{ (16\%)}\end{tabular}  & \begin{tabular}[c]{@{}r@{}}\textbf{12\%}\\\textbf{ (84\%)}\end{tabular}  & \begin{tabular}[c]{@{}r@{}}15\% \\ (100\%)\end{tabular} & \begin{tabular}[c]{@{}r@{}}4\%\\ (5\%)\end{tabular}  & \begin{tabular}[c]{@{}r@{}}\textbf{81\% }\\\textbf{ (95\%)}\end{tabular} & \begin{tabular}[c]{@{}r@{}}85\% \\ (100\%)\end{tabular}  \\ 
\cline{2-8}
                        & Hispanic                                                    & \begin{tabular}[c]{@{}r@{}}0\% \\ (0\%)\end{tabular}                     & \begin{tabular}[c]{@{}r@{}}8\% \\ (100\%)\end{tabular}                   & \begin{tabular}[c]{@{}r@{}}8\%\\ (100\%)\end{tabular}   & \begin{tabular}[c]{@{}r@{}}8\%\\ (9\%)\end{tabular}  & \begin{tabular}[c]{@{}r@{}}84\% \\ (91\%)\end{tabular}                   & \begin{tabular}[c]{@{}r@{}}92\% \\ (100\%)\end{tabular}  \\ 
\cline{2-8}
                        & Other                                                       & \begin{tabular}[c]{@{}r@{}}0\% \\ (0\%)\end{tabular}                     & \begin{tabular}[c]{@{}r@{}}20\%\\ (100\%)\end{tabular}                   & \begin{tabular}[c]{@{}r@{}}20\% \\ (100\%)\end{tabular} & \begin{tabular}[c]{@{}r@{}}4\%\\ (5\%)\end{tabular}  & \begin{tabular}[c]{@{}r@{}}76\% \\ (95\%)\end{tabular}                   & \begin{tabular}[c]{@{}r@{}}80\% \\ (100\%)\end{tabular}  \\ 
\hline
\multirow{6}{*}{Male}   & \begin{tabular}[c]{@{}r@{}}African-\\ American\end{tabular} & \begin{tabular}[c]{@{}r@{}}\textbf{15\% }\\\textbf{ (37\%)}\end{tabular} & \begin{tabular}[c]{@{}r@{}}\textbf{25\%}\\\textbf{ (63\%)}\end{tabular}  & \begin{tabular}[c]{@{}r@{}}41\% \\ (100\%)\end{tabular} & \begin{tabular}[c]{@{}r@{}}8\%\\ (14\%)\end{tabular} & \begin{tabular}[c]{@{}r@{}}\textbf{51\% }\\\textbf{ (86\%)}\end{tabular} & \begin{tabular}[c]{@{}r@{}}59\% \\ (100\%)\end{tabular}  \\ 
\cline{2-8}
                        & Asian                                                       & \begin{tabular}[c]{@{}r@{}}12\%\\ (75\%)\end{tabular}                    & \begin{tabular}[c]{@{}r@{}}4\%\\ (25\%)\end{tabular}                     & \begin{tabular}[c]{@{}r@{}}16\% \\ (100\%)\end{tabular} & \begin{tabular}[c]{@{}r@{}}0\%\\ (0\%)\end{tabular}  & \begin{tabular}[c]{@{}r@{}}84\% \\ (100\%)\end{tabular}                  & \begin{tabular}[c]{@{}r@{}}84\% \\ (100\%)\end{tabular}  \\ 
\cline{2-8}
                        & Caucasian                                                   & \begin{tabular}[c]{@{}r@{}}\textbf{5\%}\\\textbf{ (26\%)}\end{tabular}   & \begin{tabular}[c]{@{}r@{}}\textbf{14\% }\\\textbf{ (74\%)}\end{tabular} & \begin{tabular}[c]{@{}r@{}}19\%\\ (100\%)\end{tabular}  & \begin{tabular}[c]{@{}r@{}}8\%\\ (10\%)\end{tabular} & \begin{tabular}[c]{@{}r@{}}\textbf{73\%}\\\textbf{ (90\%)}\end{tabular}  & \begin{tabular}[c]{@{}r@{}}81\% \\ (100\%)\end{tabular}  \\ 
\cline{2-8}
                        & Hispanic                                                    & \begin{tabular}[c]{@{}r@{}}\textbf{3\% }\\\textbf{ (15\%)}\end{tabular}  & \begin{tabular}[c]{@{}r@{}}19\% \\ (85\%)\end{tabular}                   & \begin{tabular}[c]{@{}r@{}}22\%\\ (100\%)\end{tabular}  & \begin{tabular}[c]{@{}r@{}}7\%\\ (9\%)\end{tabular}  & \begin{tabular}[c]{@{}r@{}}71\% \\ (91\%)\end{tabular}                   & \begin{tabular}[c]{@{}r@{}}78\% \\ (100\%)\end{tabular}  \\ 
\cline{2-8}
                        & \begin{tabular}[c]{@{}r@{}}Native \\ American\end{tabular}  & \begin{tabular}[c]{@{}r@{}}14\% \\ (50\%)\end{tabular}                   & \begin{tabular}[c]{@{}r@{}}14\%\\ (50\%)\end{tabular}                    & \begin{tabular}[c]{@{}r@{}}29\% \\ (100\%)\end{tabular} & \begin{tabular}[c]{@{}r@{}}0\%\\ (0\%)\end{tabular}  & \begin{tabular}[c]{@{}r@{}}71\%\\ (100\%)\end{tabular}                   & \begin{tabular}[c]{@{}r@{}}71\% \\ (100\%)\end{tabular}  \\ 
\cline{2-8}
                        & Other                                                       & \begin{tabular}[c]{@{}r@{}}9\% \\ (45\%)\end{tabular}                    & \begin{tabular}[c]{@{}r@{}}11\% \\ (55\%)\end{tabular}                   & \begin{tabular}[c]{@{}r@{}}20\%\\ (100\%)\end{tabular}  & \begin{tabular}[c]{@{}r@{}}7\%\\ (9\%)\end{tabular}  & \begin{tabular}[c]{@{}r@{}}73\%\\ (91\%)\end{tabular}                    & \begin{tabular}[c]{@{}r@{}}80\% \\ (100\%)\end{tabular}  \\
\hline
\end{tabular}

\vspace{1em}

\begin{tabular}{r|r|rrr|rrr|}
\cline{2-8}
                                              & Actual                                                      & \multicolumn{3}{c|}{+}                                                                                                                                                                                                       & \multicolumn{3}{c|}{-}                                                                                                                                                                                                                \\ \cline{2-8} 
                                              & Predicted                                                   & \multicolumn{1}{c|}{+}                                                               & \multicolumn{1}{c|}{-}                                                      & \multicolumn{1}{c|}{Total}                              & \multicolumn{1}{c|}{+}                                                              & \multicolumn{1}{c|}{-}                                                               & \multicolumn{1}{c|}{Total}                               \\ \hline
\multicolumn{1}{|r|}{\multirow{5}{*}{Female}} & \begin{tabular}[c]{@{}r@{}}African-\\ American\end{tabular} & \multicolumn{1}{r|}{\begin{tabular}[c]{@{}r@{}}5\% \\ (40\%)\end{tabular}}           & \multicolumn{1}{r|}{\begin{tabular}[c]{@{}r@{}}7\%\\  (60\%)\end{tabular}}  & \begin{tabular}[c]{@{}r@{}}12\% \\ (100\%)\end{tabular} & \multicolumn{1}{r|}{\begin{tabular}[c]{@{}r@{}}21\%\\ (23\%)\end{tabular}}          & \multicolumn{1}{r|}{\begin{tabular}[c]{@{}r@{}}67\% \\ (100\%)\end{tabular}}         & \begin{tabular}[c]{@{}r@{}}88\%\\  (100\%)\end{tabular}  \\ \cline{2-8} 
\multicolumn{1}{|r|}{}                        & Asian                                                       & \multicolumn{1}{r|}{\begin{tabular}[c]{@{}r@{}}0\% \\ (n/a)\end{tabular}}            & \multicolumn{1}{r|}{\begin{tabular}[c]{@{}r@{}}0\% \\ (n/a)\end{tabular}}   & \begin{tabular}[c]{@{}r@{}}0\% \\ (100\%)\end{tabular}  & \multicolumn{1}{r|}{\begin{tabular}[c]{@{}r@{}}0\%\\ (0\%)\end{tabular}}            & \multicolumn{1}{r|}{\begin{tabular}[c]{@{}r@{}}100\% \\ (100\%)\end{tabular}}        & \begin{tabular}[c]{@{}r@{}}100\%\\  (100\%)\end{tabular} \\ \cline{2-8} 
\multicolumn{1}{|r|}{}                        & Caucasian                                                   & \multicolumn{1}{r|}{\textbf{\begin{tabular}[c]{@{}r@{}}2\% \\ (35\%)\end{tabular}}}  & \multicolumn{1}{r|}{\begin{tabular}[c]{@{}r@{}}5\% \\ (65\%)\end{tabular}}  & \begin{tabular}[c]{@{}r@{}}7\%\\  (100\%)\end{tabular}  & \multicolumn{1}{r|}{\textbf{\begin{tabular}[c]{@{}r@{}}12\%\\ (13\%)\end{tabular}}} & \multicolumn{1}{r|}{\textbf{\begin{tabular}[c]{@{}r@{}}81\% \\ (87\%)\end{tabular}}} & \begin{tabular}[c]{@{}r@{}}93\% \\ (100\%)\end{tabular}  \\ \cline{2-8} 
\multicolumn{1}{|r|}{}                        & Hispanic                                                    & \multicolumn{1}{r|}{\begin{tabular}[c]{@{}r@{}}0\% \\ (0\%)\end{tabular}}            & \multicolumn{1}{r|}{\begin{tabular}[c]{@{}r@{}}8\% \\ (100\%)\end{tabular}} & \begin{tabular}[c]{@{}r@{}}8\% \\ (100\%)\end{tabular}  & \multicolumn{1}{r|}{\begin{tabular}[c]{@{}r@{}}8\%\\ (9\%)\end{tabular}}            & \multicolumn{1}{r|}{\begin{tabular}[c]{@{}r@{}}84\%\\  (91\%)\end{tabular}}          & \begin{tabular}[c]{@{}r@{}}92\% \\ (100\%)\end{tabular}  \\ \cline{2-8} 
\multicolumn{1}{|r|}{}                        & Other                                                       & \multicolumn{1}{r|}{\begin{tabular}[c]{@{}r@{}}0\% \\ (0\%)\end{tabular}}            & \multicolumn{1}{r|}{\begin{tabular}[c]{@{}r@{}}4\% \\ (100\%)\end{tabular}} & \begin{tabular}[c]{@{}r@{}}4\% \\ (100\%)\end{tabular}  & \multicolumn{1}{r|}{\begin{tabular}[c]{@{}r@{}}20\%\\ (21\%)\end{tabular}}          & \multicolumn{1}{r|}{\begin{tabular}[c]{@{}r@{}}76\%\\  (79\%)\end{tabular}}          & \begin{tabular}[c]{@{}r@{}}96\% \\ (100\%)\end{tabular}  \\ \hline
\multicolumn{1}{|r|}{\multirow{6}{*}{Male}}   & \begin{tabular}[c]{@{}r@{}}African-\\ American\end{tabular} & \multicolumn{1}{r|}{\textbf{\begin{tabular}[c]{@{}r@{}}15\% \\ (65\%)\end{tabular}}} & \multicolumn{1}{r|}{\begin{tabular}[c]{@{}r@{}}8\% \\ (35\%)\end{tabular}}  & \begin{tabular}[c]{@{}r@{}}23\% \\ (100\%)\end{tabular} & \multicolumn{1}{r|}{\textbf{\begin{tabular}[c]{@{}r@{}}25\%\\ (33\%)\end{tabular}}} & \multicolumn{1}{r|}{\textbf{\begin{tabular}[c]{@{}r@{}}51\% \\ (67\%)\end{tabular}}} & \begin{tabular}[c]{@{}r@{}}77\% \\ (100\%)\end{tabular}  \\ \cline{2-8} 
\multicolumn{1}{|r|}{}                        & Asian                                                       & \multicolumn{1}{r|}{\begin{tabular}[c]{@{}r@{}}12\%\\  (100\%)\end{tabular}}         & \multicolumn{1}{r|}{\begin{tabular}[c]{@{}r@{}}0\% \\ (0\%)\end{tabular}}   & \begin{tabular}[c]{@{}r@{}}12\% \\ (100\%)\end{tabular} & \multicolumn{1}{r|}{\begin{tabular}[c]{@{}r@{}}4\%\\ (5\%)\end{tabular}}            & \multicolumn{1}{r|}{\begin{tabular}[c]{@{}r@{}}84\% \\ (96\%)\end{tabular}}          & \begin{tabular}[c]{@{}r@{}}88\% \\ (100\%)\end{tabular}  \\ \cline{2-8} 
\multicolumn{1}{|r|}{}                        & Caucasian                                                   & \multicolumn{1}{r|}{\textbf{\begin{tabular}[c]{@{}r@{}}5\% \\ (37\%)\end{tabular}}}  & \multicolumn{1}{r|}{\begin{tabular}[c]{@{}r@{}}9\% \\ (63\%)\end{tabular}}  & \begin{tabular}[c]{@{}r@{}}14\% \\ (100\%)\end{tabular} & \multicolumn{1}{r|}{\textbf{\begin{tabular}[c]{@{}r@{}}14\%\\ (16\%)\end{tabular}}} & \multicolumn{1}{r|}{\textbf{\begin{tabular}[c]{@{}r@{}}73\% \\ (84\%)\end{tabular}}} & \begin{tabular}[c]{@{}r@{}}87\% \\ (100\%)\end{tabular}  \\ \cline{2-8} 
\multicolumn{1}{|r|}{}                        & Hispanic                                                    & \multicolumn{1}{r|}{\textbf{\begin{tabular}[c]{@{}r@{}}3\% \\ (33\%)\end{tabular}}}  & \multicolumn{1}{r|}{\begin{tabular}[c]{@{}r@{}}7\% \\ (67\%)\end{tabular}}  & \begin{tabular}[c]{@{}r@{}}10\% \\ (100\%)\end{tabular} & \multicolumn{1}{r|}{\begin{tabular}[c]{@{}r@{}}19\%\\ (21\%)\end{tabular}}          & \multicolumn{1}{r|}{\begin{tabular}[c]{@{}r@{}}71\%\\  (79\%)\end{tabular}}          & \begin{tabular}[c]{@{}r@{}}90\% \\ (100\%)\end{tabular}  \\ \cline{2-8} 
\multicolumn{1}{|r|}{}                        & \begin{tabular}[c]{@{}r@{}}Native \\ American\end{tabular}  & \multicolumn{1}{r|}{\begin{tabular}[c]{@{}r@{}}14\% \\ (100\%)\end{tabular}}         & \multicolumn{1}{r|}{\begin{tabular}[c]{@{}r@{}}0\% \\ (0\%)\end{tabular}}   & \begin{tabular}[c]{@{}r@{}}14\% \\ (100\%)\end{tabular} & \multicolumn{1}{r|}{\begin{tabular}[c]{@{}r@{}}14\%\\ (17\%)\end{tabular}}          & \multicolumn{1}{r|}{\begin{tabular}[c]{@{}r@{}}71\% \\ (83\%)\end{tabular}}          & \begin{tabular}[c]{@{}r@{}}86\%\\  (100\%)\end{tabular}  \\ \cline{2-8} 
\multicolumn{1}{|r|}{}                        & Other                                                       & \multicolumn{1}{r|}{\begin{tabular}[c]{@{}r@{}}9\% \\ (55\%)\end{tabular}}           & \multicolumn{1}{r|}{\begin{tabular}[c]{@{}r@{}}7\%\\  (45\%)\end{tabular}}  & \begin{tabular}[c]{@{}r@{}}16\% \\ (100\%)\end{tabular} & \multicolumn{1}{r|}{\begin{tabular}[c]{@{}r@{}}11\%\\ (13\%)\end{tabular}}          & \multicolumn{1}{r|}{\begin{tabular}[c]{@{}r@{}}73\%\\  (87\%)\end{tabular}}          & \begin{tabular}[c]{@{}r@{}}84\% \\ (100\%)\end{tabular}  \\ \hline
\end{tabular}

\label{tab:is_ratio}
\end{table}
% Intersectional Confusion
%%%%%%%%%%%%%%%%%%%%%%%%%%%%%%%%%%%%%%%%%%%%%%%%%%%%%%%%%%%%%%%%%%

In addition to the separate analysis of race and gender, an intersectional groups analysis is performed. Produced by the two sex-based and six race-based groups, there are 12 intersectional groups. There is no observation for Native American females, so it is not included. The equal confusion test resulted in $p < 0.001$. Subsequently, the confusion parity error is computed as $\phi = 0.16$, which indicates a small but statistically significant unfairness at its upper limits with $q=10, r=4$. Tables \ref{tab:is_contingency} and \ref{tab:is_ratio} present the contingency and confusion matrices in the same fashion as the earlier analyses of sex and race. A closer analysis of the significant cells results in the following observations.

\begin{itemize}
    \item Among predicted risky African-American males, $37\%$ are actually recidivists, while the same figure drops to $26\%$ for Caucasian males, $15\%$ for Hispanic males, and $16\%$ for Caucasian females. Precision is highest for African-American males and lowest for Hispanic males and Caucasian females. Hence, the last two groups are more likely to be incorrectly predicted as risky than African-American males.
    
    \item Among predicted non-risky Caucasian females, $95\%$ are actual non-recidivists, while the same figure goes down to $90\%$ for Caucasian males and $86\%$ for African-American males. Negative predictive value is lower for African-American males. Hence, they are more likely to benefit from false negatives.
    
    \item Among actual recidivist Caucasian females, Caucasian males, and Hispanic males, only $35\%$, $37\%$, and $33\%$, respectively, are correctly predicted as risky, while the same figure is $65\%$ for African-American males. Recall is higher for African-American males. Hence, the other groups are more likely to benefit from the under-identification of risky status.
    
    \item Among actual non-recidivist Caucasian females and males, $87\%$ and $84\%$, respectively, are correctly predicted as non-risky, while the same figure drops to $67\%$ for African-American males. Specificity is lower for African-Americans. Hence, African-Americans are more likely to suffer from an under-identification of non-risky status.
\end{itemize}

The first two findings indicate an advantageous position for African-American males and disadvantageous positions for Caucasian females. Compared to the earlier findings from the non-intersectional analysis, it can be argued that the disadvantages of African-Americans lie with its male members. Similarly, the advantages of Caucasians lie more with their female members. The last two findings indicate disadvantageous positions for Caucasian males and females at comparable levels. However, the gender gap remains among African-Americans, where its male members are in a statistically significantly disadvantageous position while its female members are not.

\subsection{Remarks}

This case study demonstrates the proposed equal confusion test, confusion parity error, and post hoc fairness analysis. Unfairness in the system is successfully detected and quantified. Despite the relatively small strength of the unfairness, the post hoc analysis revealed several statistically significant fairness issues between certain groups. It also showed that a group that bears negative impacts from one perspective might be a beneficiary from another, indicating that fairness implications are not necessarily straightforward. Furthermore, the intersectional group analysis enabled the mapping of observed unfairness to more refined subgroups. Overall, it may be concluded that using COMPAS in critical criminal justice decisions is worrisome.

There are two main limitations of this case study. First, only the data from a particular county in Florida is available, making the findings' generalizability questionable. Second, the number of cases is limited, particularly for certain races and many intersectional groups, hindering the possibility of obtaining statistically significant results for those groups.

\section{Conclusion}\label{sec:con}
Given the larger impact AI has started to have on human lives, the need for accountability of automated decision systems has become inevitable. Fairness is critical to such accountability efforts, improving the trust placed in AI systems to reap technological benefits without causing harm. However, there is an abundance of fairness metrics that are well refined, often incompatible, and subject to "cherry-picking." Therefore, the need for a unifying fairness assessment methodology is paramount.

The main contributions of this study are the proposed equal confusion test, the confusion parity error, and the associated methodology for post hoc fairness analysis. 
The equal confusion test checks whether the system exhibits any unfair behavior. If unfairness is detected, the confusion parity error is utilized to quantify the magnitude of unfairness. A table is provided to interpret the values of the confusion parity error. 
Finally, the post hoc analysis is employed to examine the characteristics and positively/negatively impacted groups via identifying confusion matrix cells with statistically significant divergence from their expected values.

The use of the proposed test, metric and post hoc analysis methods are demonstrated via a case study. The case study employs real-world data from COMPAS, a criminal risk assessment tool used in the US to assist pretrial release decisions. The findings indicate that COMPAS is not fair, and discrepancies exist between different sex and race groups and intersectional groups. Specifically, African-American males and Caucasians show divergent behavior that is statistically significant. From some perspectives, one group is at a disadvantage while the same group is at an advantage from other perspectives. The findings indicate that the use of COMPAS in critical decisions is problematic.

The foreseen future research is two-fold. First, analogous tests, measures, and post hoc analyses can be developed for regression tasks. Second, the proposed methodology can be employed to assess group fairness in various currently deployed automated decision systems. %Third, the proposed methodology can be adopted to assess fairness as part of a larger accountability framework.

\section*{Acknowledgment}
This material is based upon work supported by the National Science Foundation under Grant CCF-2131504. Any opinions, findings, and conclusions or recommendations expressed in this material are those of the authors and do not necessarily reflect the views of the National Science Foundation.


\begin{thebibliography}{00}
\bibitem{angwinMachineBias} J. Angwin, J. Larson, S. Mattu, and L. Kirchner, “Machine bias.” May 23, 2016. Accessed: May 19, 2022. [Online]. Available: https://www.propublica.org/article/machine-bias-risk-assessments-in-criminal-sentencing
\bibitem{vanbekkumDigitalWelfareFraud2021} M. van Bekkum and F. Z. Borgesius, “Digital welfare fraud detection and the Dutch SyRI judgment,” European Journal of Social Security, vol. 23, no. 4, pp. 323–340, Dec. 2021, doi: 10.1177/13882627211031257.
\bibitem{castelvecchiFacialRecognitionToo2020} D. Castelvecchi, “Is facial recognition too biased to be let loose?,” Nature, vol. 587, no. 7834, pp. 347–349, Nov. 2020, doi: 10.1038/d41586-020-03186-4.
\bibitem{raghavanMitigatingBiasAlgorithmic2020} M. Raghavan, S. Barocas, J. Kleinberg, and K. Levy, “Mitigating bias in algorithmic hiring: evaluating claims and practices,” in Proc. Conference on Fairness, Accountability, and Transparency, New York, NY, Jan. 27, 2020, pp. 469–481. doi: 10.1145/3351095.3372828.
\bibitem{mcleanDigitalJusticeAustralian2019} J. McLean and R. Mackenzie, “Digital justice in Australian visa application processes?,” Alternative Law Journal, vol. 44, no. 4, pp. 291–296, Dec. 2019, doi: 10.1177/1037969X19853685.
\bibitem{choudhuryRoleArtificialIntelligence2020} A. Choudhury and O. Asan, “Role of artificial intelligence in patient safety outcomes: systematic literature review,” JMIR Medical Informatics, vol. 8, no. 7, p. e18599, Jul. 2020, doi: 10.2196/18599.
\bibitem{gursoySystemCardsAIBased2022} F. Gursoy and I. A. Kakadiaris, “System cards for AI-based decision-making for public policy.” arXiv, Mar. 01, 2022. doi: 10.48550/arXiv.2203.04754.
\bibitem{clarkeAlgorithmicAccountabilityAct2022} Y. D. Clarke, Algorithmic Accountability Act of 2022. 2022. Accessed: May 19, 2022. [Online]. Available: https://www.congress.gov/bill/117th-congress/house-bill/6580
\bibitem{EURLex52021PC0206EURLex} {European Commission}, Artificial Intelligence Act. 2021. Accessed: May 19, 2022. [Online]. Available: https://eur-lex.europa.eu/legal-content/EN/TXT/?uri=CELEX\%3A52021PC0206
\bibitem{GuidanceAIAuditing} Information Commissioner’s Office, “Guidance on the AI auditing framework: draft guidance for consultation.” Information Commissioner’s Office, Feb. 2020. Accessed: Jan. 07, 2022. [Online]. Available: https://ico.org.uk/about-the-ico/ico-and-stakeholder-consultations/ico-consultation-on-the-draft-ai-auditing-framework-guidance-for-organisations/
\bibitem{mehrabiSurveyBiasFairness2021} N. Mehrabi, F. Morstatter, N. Saxena, K. Lerman, and A. Galstyan, “A survey on bias and fairness in machine learning,” ACM Comput. Surv., vol. 54, no. 6, p. 115:1-115:35, Jul. 2021, doi: 10.1145/3457607.
\bibitem{samoraniOverbookedOverlookedMachine2021} M. Samorani, S. L. Harris, L. G. Blount, H. Lu, and M. A. Santoro, “Overbooked and overlooked: machine learning and racial bias in medical appointment scheduling,” Manufacturing \& Service Operations Management, Aug. 2021, doi: 10.1287/msom.2021.0999.
\bibitem{pratesAssessingGenderBias2020} M. O. R. Prates, P. H. Avelar, and L. C. Lamb, “Assessing gender bias in machine translation: a case study with Google Translate,” Neural Computing and Applications, vol. 32, no. 10, pp. 6363–6381, May 2020, doi: 10.1007/s00521-019-04144-6.
\bibitem{chuDigitalAgeismChallenges2022} C. H. Chu, R. Nyrup, K. Leslie, J. Shi, A. Bianchi, A. Lyn, M. McNicholl, S. Khan, S. Rahimi, and A. Grenier, “Digital ageism: challenges and opportunities in artificial intelligence for older adults,” The Gerontologist, Jan. 2022, doi: 10.1093/geront/gnab167.
\bibitem{whittakerDisabilityBiasAI} M. Whittaker, M. Alper, O. College, L. Kaziunas, and M. R. Morris, “Disability, bias, and AI.” Nov. 2019. Accessed: May 19, 2022. [Online]. Available: https://ainowinstitute.org/disabilitybiasai-2019.pdf
\bibitem{petersAlgorithmicPoliticalBias2022} U. Peters, “Algorithmic political bias in artificial intelligence systems,” Philosophy \& Technology, vol. 35, no. 2, p. 25, Mar. 2022, doi: 10.1007/s13347-022-00512-8.
\bibitem{abidPersistentAntiMuslimBias2021} A. Abid, M. Farooqi, and J. Zou, “Persistent anti-Muslim bias in large language models,” in Proc. AAAI/ACM Conference on AI, Ethics, and Society, New York, NY, Jul. 21, 2021, pp. 298–306. doi: 10.1145/3461702.3462624.
\bibitem{makhloufApplicabilityMachineLearning2021} K. Makhlouf, S. Zhioua, and C. Palamidessi, “On the applicability of machine learning fairness notions,” ACM SIGKDD Explorations Newsletter, vol. 23, no. 1, pp. 14–23, May 2021, doi: 10.1145/3468507.3468511.
\bibitem{castelnovoClarificationNuancesFairness2022} A. Castelnovo, R. Crupi, G. Greco, D. Regoli, I. G. Penco, and A. C. Cosentini, “A clarification of the nuances in the fairness metrics landscape,” Scientific Reports, vol. 12, no. 1, p. 4209, Mar. 2022, doi: 10.1038/s41598-022-07939-1.
\bibitem{vermaFairnessDefinitionsExplained2018} S. Verma and J. Rubin, “Fairness definitions explained,” in Proc. International Workshop on Software Fairness, New York, NY, May 29, 2018, pp. 1–7. doi: 10.1145/3194770.3194776.
\bibitem{bellamyAIFairness3602019} R. K. E. Bellamy, K. Dey, M. Hind, S. C. Hoffman, S. Houde, K. Kannan, P. Lohia, J. Martino, S. Mehta, A. Mojsilović, S. Nagar, K. N. Ramamurthy, J. Richards, D. Saha, P. Sattigeri, M. Singh, K. R. Varshney, and Y. Zhang, “AI Fairness 360: an extensible toolkit for detecting and mitigating algorithmic bias,” IBM Journal of Research and Development, vol. 63, no. 4/5, p. 4:1-4:15, Jul. 2019, doi: 10.1147/JRD.2019.2942287.
\bibitem{segalFairnessEyesData2021} S. Segal, Y. Adi, B. Pinkas, C. Baum, C. Ganesh, and J. Keshet, “Fairness in the eyes of the data: certifying machine-learning models,” in Proc. AAAI/ACM Conference on AI, Ethics, and Society, New York, NY, Jul. 21, 2021, pp. 926–935. doi: 10.1145/3461702.3462554.
\bibitem{chouldechovaSnapshotFrontiersFairness2020} A. Chouldechova and A. Roth, “A snapshot of the frontiers of fairness in machine learning,” Communications of the ACM, vol. 63, no. 5, pp. 82–89, Apr. 2020, doi: 10.1145/3376898.
\bibitem{razGroupFairnessIndependence2021} T. Räz, “Group fairness: independence revisited,” in Proc. ACM Conference on Fairness, Accountability, and Transparency, New York, NY, Mar. 3, 2021, pp. 129–137. doi: 10.1145/3442188.3445876.
\bibitem{berkFairnessCriminalJustice2021} R. Berk, H. Heidari, S. Jabbari, M. Kearns, and A. Roth, “Fairness in criminal justice risk assessments: the state of the art,” Sociological Methods \& Research, vol. 50, no. 1, pp. 3–44, 2021, doi: 10.1177/0049124118782533.
\bibitem{barocasFairnessMachineLearning2019} S. Barocas, M. Hardt, and A. Narayanan, Fairness and Machine Learning. fairmlbook.org, 2019.
\bibitem{fouldsIntersectionalDefinitionFairness2020} J. R. Foulds, R. Islam, K. Keya, and S. Pan, “An intersectional definition of fairness,” in Proc. International Conference on Data Engineering, Los Alamitos, CA, Apr. 2020, pp. 1918–1921. doi: 10.1109/ICDE48307.2020.00203.
\bibitem{kearnsPreventingFairnessGerrymandering2018} M. Kearns, S. Neel, A. Roth, and Z. S. Wu, “Preventing fairness gerrymandering: auditing and learning for subgroup fairness,” in Proc. International Conference on Machine Learning, Jul. 3, 2018, pp. 2564–2572. [Online]. Available: https://proceedings.mlr.press/v80/kearns18a.html
\bibitem{cramerMathematicalMethodsStatistics1946} H. Cramer, Mathematical Methods of Statistics. Princeton: Princeton University Press, 1946.
\bibitem{matthewsComparisonPredictedObserved1975} B. W. Matthews, “Comparison of the predicted and observed secondary structure of T4 phage lysozyme,” Biochimica et Biophysica Acta (BBA) - Protein Structure, vol. 405, no. 2, pp. 442–451, Oct. 1975, doi: 10.1016/0005-2795(75)90109-9.
\bibitem{cohenStatisticalPowerAnalysis} J. Cohen, Statistical Power Analysis for the Behavioral Sciences, 2nd ed. Routledge, 1988.
\bibitem{fergusonEffectSizePrimer2009} C. J. Ferguson, “An effect size primer: a guide for clinicians and researchers,” Professional Psychology: Research and Practice, vol. 40, no. 5, pp. 532–538, 2009, doi: 10.1037/a0015808.
\bibitem{Sharpe2015YourCT} D. M. Sharpe, “Your chi-square test is statistically significant: now what?,” Practical Assessment, Research and Evaluation, vol. 20, no. 8, pp. 1–10, 2015.
\bibitem{habermanAnalysisResidualsCrossClassified1973} S. J. Haberman, “The analysis of residuals in cross-classified tables,” Biometrics, vol. 29, no. 1, pp. 205–220, 1973, doi: 10.2307/2529686.
\bibitem{Agresti2007} A. Agresti, An Introduction to Categorical Data Analysis, 2nd ed. New York: Wiley-Interscience, 2007.
\bibitem{macdonaldTypeErrorRate2000} P. L. MacDonald and R. C. Gardner, “Type I error rate comparisons of post hoc procedures for I x j chi-square tables,” Educational and Psychological Measurement, vol. 60, no. 5, pp. 735–754, Oct. 2000, doi: 10.1177/00131640021970871.
\bibitem{cochranMethodsStrengtheningCommon1954} W. G. Cochran, “Some methods for strengthening the common x² tests,” Biometrics, vol. 10, pp. 417–451, 1954, doi: 10.2307/3001616.
\bibitem{coxPostHocPairWise1993} M. K. Cox and C. H. Key, “Post hoc pair-wise comparisons for the chi-square test of homogeneity of proportions,” Educational and Psychological Measurement, vol. 53, no. 4, pp. 951–962, Dec. 1993, doi: 10.1177/0013164493053004008.
\bibitem{kirkpatrickItNotAlgorithm2017} K. Kirkpatrick, “It’s not the algorithm, it’s the data,” Communications of the ACM, vol. 60, no. 2, pp. 21–23, Jan. 2017, doi: 10.1145/3022181.
\bibitem{mattuHowWeAnalyzed} J. Larson, S. Mattu, L. Kirchner, and J. Angwin, “How we analyzed the COMPAS recidivism algorithm.” Accessed: May 19, 2022. [Online]. Available: https://www.propublica.org/article/how-we-analyzed-the-compas-recidivism-algorithm
\bibitem{israniAlgorithmicDueProcess} E. Israni and E. Chang, “Algorithmic due process: mistaken accountability and attribution in State v. Loomis.” Aug. 31, 2017. Accessed: Apr. 14, 2022. [Online]. Available: https://jolt.law.harvard.edu/digest/algorithmic-due-process-mistaken-accountability-and-attribution-in-state-v-loomis-1
\bibitem{n.w.2d749StateLoomis} Harvard Law Review, “State v. Loomis.” Mar. 10, 2017. Accessed: May 19, 2022. [Online]. Available: https://harvardlawreview.org/2017/03/state-v-loomis/
\bibitem{LoomisWisconsin} “Loomis v. Wisconsin.” Jun. 26, 2017. Accessed: May 19, 2022. [Online]. Available: https://www.scotusblog.com/case-files/cases/loomis-v-wisconsin/
\bibitem{dieterichCOMPASRiskScales2016} W. Dieterich, C. Mendoza, and T. Brennan, “COMPAS risk scales: demonstrating accuracy equity and predictive parity.” Jul. 08, 2016. Accessed: May 19, 2022. [Online]. Available: https://go.volarisgroup.com/rs/430-MBX-989/images/ProPublica\_Commentary\_Final\_070616.pdf
\bibitem{washingtonHowArgueAlgorithm2019} A. L. Washington, “How to argue with an algorithm: lessons from the COMPAS ProPublica debate,” The Colorado Technology Law Journal, vol. 17, no. 1, pp. 131–160, Apr. 2019.
\bibitem{jacksonSettingRecordStraight2020} E. Jackson and C. Mendoza, “Setting the record straight: what the COMPAS core risk and need assessment is and is not.” Mar. 31, 2020. Accessed: May 19, 2022. [Online]. Available: https://hdsr.mitpress.mit.edu/pub/hzwo7ax4/
\bibitem{propublicaDataAnalysisMachine2022} ProPublica, “Data and analysis for ‘machine bias.’” May 20, 2022. Accessed: May 19, 2022. [Online]. Available: https://github.com/propublica/compas-analysis
\bibitem{ViolentCrime} FBI, “Violent crime.” Accessed: Apr. 12, 2022. [Online]. Available: https://ucr.fbi.gov/crime-in-the-u.s/2010/crime-in-the-u.s.-2010/violent-crime/violent-crime
\bibitem{northpointePractitionersGuideCOMPAS2012} Northpointe, “Practitioners guide to COMPAS.” 2012. Accessed: Apr. 14, 2022. [Online]. Available: https://njoselson.github.io/pdfs/FieldGuide2\_081412.pdf




\end{thebibliography}
\end{document}